# Why does my medical AI look at pictures of birds? Exploring the efficacy of transfer learning across domain boundaries

**Frederic Jonske (M.Sc.)**[1], Moon Kim (Dr. med.)[1], Enrico Nasca (M.Sc.)[1], Janis Evers (M.Sc.)[1], Johannes Haubold (Dr. med.)[2], René Hosch (Dr. rer. med.)[1,2], Felix Nensa (Prof.)[1,2], Michael Kamp (PhD)[1,3,4], Constantin Seibold (M.Sc.)[1], Jan Egger (Prof.)[1,5], Jens Kleesiek (Prof.)[1,5,6]

{**frederic.jonske**, moon.kim, enrico.nasca, johannes.haubold, rene.hosch, felix.nensa, michael.kamp, constantin.seibold, jan.egger, jens.kleesiek}@uk-essen.de
janis.evers@stud.uni-due.de

1 - Institute of AI in Medicine (IKIM), University Medicine Essen (AöR), University Duisburg-Essen, Germany
(Address: Institut für künstliche Intelligenz in der Medizin (IKIM), Girardetstr. 2, 45131 Essen, NRW, Germany)
2 - Institute of Diagnostic and Interventional Radiology and Neuroradiology,
University Medicine Essen (AöR), Germany
3 - Institute for Neuroinformatics, Ruhr University Bochum, Germany
4 - Department of Data Science & AI, Monash University, Australia
5 - Cancer Research Center Cologne Essen (CCCE), University Medicine Essen (AöR), Germany
6 - Department of Physics, TU Dortmund University, Dortmund, Germany

**Corresponding author can be reached at:**

- E-mail: frederic.jonske@uk-essen.de
- Phone: +49 1575 4095555
- Address: Hohenstein 120, 42283 Wuppertal, NRW, Germany

## Abstract

**Purpose:** In medical deep learning, models not trained from scratch are typically fine-tuned based on ImageNet-pretrained models. We posit that pretraining on data from the domain of the downstream task should almost always be preferable.
**Materials and methods:** We leverage RadNet-12M and RadNet-1.28M, datasets containing more than 12 million/1.28 million acquired CT image slices from 90,663 individual scans, and explore the efficacy of self-supervised, contrastive pretraining on the medical and natural image domains. We compare the respective performance gains for five downstream tasks. For each experiment, we report accuracy, AUC, or DICE score and uncertainty estimations based on four separate runs. We quantify significance using Welch's t-test. Finally, we perform feature space analysis to characterize the nature of the observed performance gains.
**Results:** We observe that intra-domain transfer (RadNet pretraining and CT-based tasks) compares favorably to cross-domain transfer (ImageNet pretraining and CT-based tasks), generally achieving comparable or improved performance – $\Delta = +0.44\%$ ($p = 0.541$) when fine-tuned on RadNet-1.28M, $\Delta = +2.07\%$ ($p = 0.025$) when linearly evaluating on RadNet-1.28M, and $\Delta = +1.63\%$ ($p = 0.114$) when fine-tuning on 1% of RadNet-1.28M data. This intra-domain advantage extends to LiTS 2017, another CT-based dataset, but not to other medical imaging modalities. A corresponding intra-domain advantage was also observed for natural images. Outside the CT image domain, ImageNet-pretrained models generalized better than RadNet-pretrained models.

We further demonstrate that pretraining on medical images yields domain-specific features that are preserved during fine-tuning, and which correspond to macroscopic image properties and structures.

**Conclusion:** We conclude that intra-domain pretraining generally outperforms cross-domain pretraining, but that very narrow domain definitions apply. Put simply, pretraining on CT images instead of natural images yields an advantage when fine-tuning on CT images, and only on CT images. We further conclude that ImageNet pretraining remains a strong baseline, as well as the best choice for pretraining if only insufficient data from the target domain is available. Finally, we publish our pretrained models and pretraining guidelines as a baseline for future research.

## 1 Introduction

Is there a guaranteed-to-be-best dataset to pretrain on, given a task? One could argue that any task's dataset is naturally best represented by features learned during pretraining on itself, or on a dataset from the same underlying distribution. One could alternatively argue that any image should be representable by a generic set of features, no matter where that set is learned. After all, humans also perform transfer learning by first being "pretrained" on natural images while growing up, before "getting fine-tuned" on medical images to become radiologists. In this paper, we will systematically explore this question regarding medical and natural image domains, using contrastive, self-supervised pretraining.

In the literature, several works explore transfer learning applications from medical pretraining to medical downstream tasks. Some have reported that pretraining on one domain improves fine-tuning on another. Others, however, have found that unsupervised pretraining offered no substantial improvements. This implies that both lines of reasoning provided by us have at least some merits.

Cherti & Jitsev [1] explored the scalability of supervised pretraining and transfer learning in medical-medical, natural-natural and cross-domain scenarios. Their findings indicated that in some cases, upscaling the available pretraining data led to proportionally increased performance, while for few-shot tasks the effect disappeared. They additionally reported that models pretrained on ImageNet-21k [2] outperformed models pretrained on the largest available X-ray datasets when fine-tuning on X-ray image downstream tasks. Notably, the combined number of labeled X-rays available for this comparison was about 16 times less than the images contained in ImageNet-21k (approx. 873k vs. approx. 14M).

Leveraging around 100 million medical images from multiple modalities for self-supervised pretraining, Ghesu et al. reported that their self-supervised algorithm, together with the significant dataset upscaling, yielded a significant advantage over both training from scratch and pretraining with other methods [3].

Mei et al. created RadImageNet [4], a dataset composed of 1.4 million medical images. They reported that models trained on RadImageNet achieved superior performance on medical tasks, compared to ImageNet-pretrained models. They further observed a correlation between the size of the dataset in the downstream task and the transfer learning performance, noting that RadImageNet yielded greater improvements on smaller datasets.

Azizi et al. created REMEDIS [5], a multi-supervision level approach for transfer learning on medical tasks. REMEDIS combines large-scale supervised learning on natural images with self-supervised learning on medical images, and achieves significant performance gains on 15 medical tasks, compared to supervised baselines.

While noting that ImageNet-pretraining was often a *de facto* method for medical image classification problems, Raghu et al. observed across a range of different model architectures and medical imaging tasks, that pretraining with ImageNet yielded no significant performance boost and that meaningful feature reuse did not always occur after pretraining [6].

Newell & Deng experimentally probed the efficacy of self-supervised training using synthetic data and various pretraining algorithms [7]. They observed an anti-proportional correlation between pretraining utility and the number fine-tuning images and noted that in some cases, pretraining offered only accelerated convergence or even no performance gain.

In a study on cross-domain generalization [8], Cohen et al. found that many of their models could perform well after fine-tuning on the same downstream task, yet disagree in their predictions. Moreover, models that were in strong agreement nonetheless performed poorly, implying that generalization to other tasks is often a nuanced process.
A study by Gururangan et al. [9] explored intra-domain transfer learning on Natural Language Processing (NLP) tasks, using multi-phased pretraining on different datasets. They found that spending at least some time training on data from the domain of the downstream task led to consistent performance gains.

Based on these previous findings and our initial lines of reasoning, we formulate the following hypotheses:
**I)** When applying transfer learning, any difference between the pretraining and fine-tuning task domains produces a *generalization gap* [10]. This occurs because the two underlying data distributions, and consequently learned feature subsets, are likely to be different at least to some degree [11], [12]. Intuitively, we expect this gap to shrink or disappear in an intra-domain transfer learning scenario, since the data distributions are highly similar or identical. In essence, we propose that **"there is no better data to train medical questions on than medical data"**.
**II)** The generalization gap effect is not symmetrical. A model trained on a specific pretraining dataset may learn a superset of the features it would learn from a different dataset. While this is most likely not immediately true for every downstream task, it may well be true for a significant fraction of them. We posit that **"some pretraining datasets are principally more useful than others"**.
Investigating these hypotheses not only provides fundamental insights into the nature of fine-tuning, domain adaptation and transfer learning, but will, more importantly, yield practical strategies to improve our medical AI models.

Our main contributions are as follows:
- To validate these hypotheses, we leverage RadNet-1.28M and RadNet-12M, two pretraining datasets comprising over 12 million CT image slices from our local hospital, the largest dataset on which this kind of analysis has been performed to date. We also leverage four publicly available datasets from the natural and medical image domain.
- We design an experimental setup (cf. Fig. 1 and the Methods section) to compare unsupervised pretraining efficacy depending on the fine-tuning dataset domain. We additionally extend previous work by exploring the dependency of the downstream performance on task complexity, and in a linear evaluation vs fine-tuning scenario.
- We present our results and uncertainty estimations, discuss them with respect to our hypotheses, and perform novel feature space analysis for our models. Here, we demonstrate the existence of unique, explainable image features learned only during intra-domain pretraining that are preserved during fine-tuning.
- Finally, we discuss practical guidelines for the choice of pretraining datasets, depending on downstream tasks, and publish our code framework and all pretrained/fine-tuned models as a baseline for future research or the extension of our work.

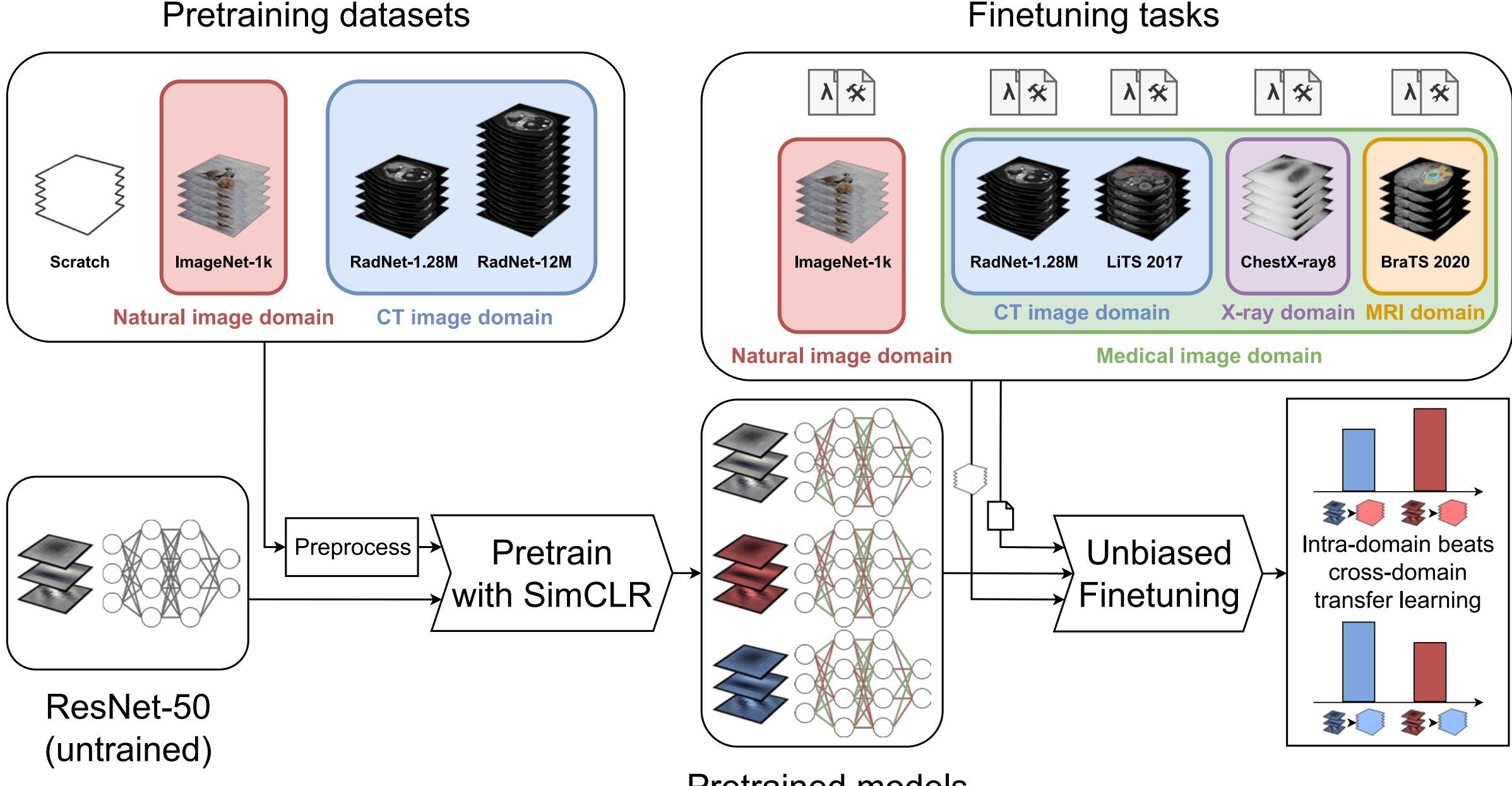

**Fig. 1: Schematic visualization of the transfer learning experiment.**
During the transfer learning experiment, models are pretrained on medical, natural, or no images (denoted by five stacked images drawn from the actual dataset, top left). The value of the pretraining in terms of performance gains is measured across several downstream tasks (top right), which also come from the medical or natural image domain. Each downstream task has a hyperparameter set and build instructions (denoted by the λ and 🛠 symbols) that are shared between executions. This makes extending the framework straightforward and more easily reproducible.

## 2 Methods

For the purposes of this paper, we define a domain as the collection of all data points that satisfy some condition. For example, a very general domain would be the domain of medical images, containing all images with any medical context. A smaller, more distinct domain would be the CT image domain, or the domain of abdominal CTs. Each of the datasets below comes from one of these domains.

### 2.1 Datasets and tasks

#### 2.1.1 RadNet-X

The RadNet-X datasets are composed purely of CT slices. We opted for a single modality because we intend to test our hypotheses on a narrow, clearly defined domain. The CT data stems from the University Hospital Essen, Germany, and was retrospectively collected for this study. The criteria are as follows: There were no exclusion criteria based on patients or dates. Scans with fewer than 30 slices in the z-direction were excluded to filter for high resolution scans. Slices with <25% coverage were removed. This process yielded 12,034,617 slices (approx. 84.8% the size of ImageNet-21k) from 90,663 CTs. Data used by Koitka et al. [13] and Jonske et al. [14] was used in this study. https://github.com/UMEssen/DicomDeidentify was used to deidentify all data. The slices were automatically labeled based on metadata, dividing them into 7 anatomical classes – “head”, “thorax”, “abdomen”, “extremities”, “pelvis”, “spine”, and “other”. From these 12 million slices, two datasets were created:

- The larger dataset, called RadNet-12M, contains all 12 million data. It is used as one of the pretraining options.
- RadNet-1.28M contains 1.28 million data and is thus the size of ImageNet-1k. It is a subset of RadNet-12M. RadNet-1.28M maintains the same class ratio as RadNet-12M, except for instances of the "other" class, which are excluded to improve fine-tuning stability. RadNet-1.28M is used both as a pretraining and as a fine-tuning dataset in our experiments.

In both datasets, slices from the same volume can only appear in either the train, validation, or test set, but not in multiple ones.

Due to human anatomy, neighboring CT slices are correlated, effectively reducing the amount of information contained in the dataset compared to one in which every image slice comes from a different scan. To facilitate a fair comparison with ImageNet, we offset this effect by interpolating between the results for RadNet-1.28M and -12M, thus simulating a larger dataset. An upper and lower bound for the factor by which RadNet-1.28M must be upscaled are calculated (see supplementary materials) as $S_{LB} = 1.136$ and $S_{UB} = 3.807$. As discussed in the supplementary materials, we consider RadNet-1.28M (UB-adjusted) as the closest radiological imaging analog to ImageNet-1k.

While we cannot make the dataset itself publicly available, the pretrained model weights for both RadNet-X datasets are made publicly available (consent waived and permission granted by the responsible IRB, cn. *20-9745-BO*).

#### 2.1.2 ImageNet-1k

ImageNet-1k [2] consists of 1.28 million RGB images of 1000 classes. The associated data domain for these images is called "natural images". Human-created labels exist for every point of data and have a one-to-one relation with images. We make an even, stratified split of the official validation set to obtain a validation and test set.

#### 2.1.3 LiTS 2017

LiTS (Liver Tumor Segmentation) 2017 [15] is a segmentation dataset composed of abdominal CT images. The data comes from 131 distinct patients and totals 41'561 individual 2D slices, all of which are professionally segmented into background, liver, and tumor regions. The official validation set is evenly split to obtain a validation and test set. Slices from the same scan do not appear in different sets.

#### 2.1.4 BraTS 2020

BraTS (Brain Tumor Segmentation) 2020 [16] is a segmentation dataset composed of cranial MRT scans from 369 glioma patients. For each patient, T1, T1ce, T2, and FLAIR modalities are available. Tumor regions are divided into Tumor Core (TC), Active Tumor (AT), and Whole Tumor (WT). We remove empty slices from the dataset. We use 30 patients each from the official training dataset to construct a validation and test set. Slices from the same scan do not appear in different sets.

### 2.1.5 ChestX-Ray8

ChestX-Ray8 [17] is a medical imaging dataset containing 108,948 frontal-view X-ray images from 32,717 unique patients. Classification labels for each X-ray were automatically acquired via text mining. We construct a training and validation set from the official "train_val" set using a 90-10 split. To avoid bias from an uneven split, training and validation sets are randomized between runs of an experiment.

## 2.2 Experimental setup

To test our hypotheses, we combine all four pretraining options (training from scratch, ImageNet-1k, RadNet-1.28M, and RadNet-12M) and all five fine-tuning tasks (ImageNet classification, RadNet-1.28M classification, ChestX-Ray8 multi-label classification, LiTS 2017 segmentation, and BraTS 2020 segmentation) in several different settings.

For implementation details, consult the supplementary materials or our documentation at https://github.com/TIO-IKIM/Transfer-learning-across-domain-boundaries.

### 2.2.1 Experiment 1 (Task Dependency)

According to Hypothesis I, we expect to see a generalization gap when we the pretraining and fine-tuning dataset domains are different. If Hypothesis I is correct, we expect RadNet-pretrained models to outperform ImageNet-pretrained models on the RadNet and LiTS 2017 downstream tasks. Simultaneously, we test Hypothesis II - if the relative performance loss in the cross-domain transfer scenarios is asymmetrical, this is a strong indication that ImageNet is generally better suited for transfer learning than RadNet.

### 2.2.2 Experiment 2 (Fine-tuning vs. Linear Evaluation)

During regular fine-tuning, such as in Experiment 1, all parameters of a pretrained model are learnable and will be adjusted. When all parameters except those of the last layer of the model are frozen (cannot be updated), that training process is termed "linear evaluation". This effectively increases the difficulty of the task and enforces the reuse of pretrained features. For segmentation tasks, the added U-Net decoder is randomly initialized at the start of the linear evaluation and thus needs to be trained to achieve any meaningful results. In this case, we allow training of the weights of half the model. The expectation according to Hypothesis I is that any observed intra-domain advantage from Experiment 1 becomes more pronounced, because the model is forced to rely entirely on the features learned during pretraining.

### 2.2.3 Experiment 3 (Task Complexity)

We define the complexity of a task as the ratio between number of labeled examples and number of target classes. We increase this complexity by decreasing the training dataset size to 50%, 10%, 1%, or 0.1% in a stratified manner, or we decrease it by removing classes. According to Hypothesis I, we expect that in this low-data scenario, pretraining features learned on the task's domain should be significantly more useful than those learned in a different domain.

### 2.3 Reported metrics

For classification tasks, we report the accuracy (ImageNet, RadNet) and average AUC (ChestX-Ray8), while for segmentation tasks (LiTS 2017, BraTS 2020) we report DICE scores per class and per slice. We provide estimates of the aleatoric uncertainty of our results based on four separate runs of the same experiment, each using different seeds. Finally, we report relative performance comparisons and their significance based on Welch's t-test [18].

## 3 Results

Firstly, we observe that pretraining on any dataset yields some degree of improved performance in almost every fine-tuning scenario we tested (cf. Figs. 2 and 3, or Supplementary Tables ST1-3). We further observe that intra-domain transfer learning (e.g. RadNet→RadNet) generally offers equal or slightly improved performance compared to cross-domain transfer learning (e.g. ImageNet→RadNet), although this is not unambiguously observed everywhere. If the complexity of the task is increased by reducing the number of images per class during fine-tuning, or linear evaluation is performed, this intra-domain advantage grows substantially, both in absolute terms and in statistical significance. The relative performance of each pairing of pretraining datasets, as well as the corresponding significance can be found in the supplementary materials (Figs. SF1-3).

We further note that the observed intra-domain advantage is not symmetrical - for example, pretraining on RadNet (abbreviated *R-1.28M*) instead of ImageNet (abbreviated *I-1k*), followed by linear evaluation on RadNet, offers a statistically less significant and smaller relative performance gain than the reverse case ($\Delta_{R-1.28M} = +0.44\%$, $p = 0.541$ vs. $\Delta_{I-1k} = +0.79\%$, $p = 0.231$). This asymmetry appears to intensify substantially during linear evaluation ($\Delta_{R-1.28M\ (lineval)} = +2.07\%$, $p = 0.025$ vs. $\Delta_{I-1k\ (lineval)} = +42.92\%$, $p < 0.001$) or for increasing task complexity ($\Delta_{R-1.28M\ (1\%)} = +1.63\%$, $p = 0.114$ vs. $\Delta_{I-1k(1\%)} = +131.62\%$, $p < 0.001$). Because of this asymmetry, the ImageNet-pretrained model displays a competitive performance in most experiments.

Notably, the observed intra-domain advantage does not apply across all medical imaging tasks - RadNet-1.28M pretraining yields slightly superior or comparable performance only on RadNet and LiTS ($\Delta_{R-1.28M} = +0.30\%$, $p = 0.320$, $\Delta_{R-1.28M} = +0.53\%$, $p = 0.119$) - both CT imaging datasets - but inferior performance on BraTS, an MRI dataset, for example.

Additionally, we note that even between target classes in the same task, different behaviors can be observed. For example, RadNet-pretrained models slightly outperformed ImageNet-pretrained models in liver segmentation, but not tumor segmentation, where RadNet-1.28M (UB-adjusted) performs similarly to ImageNet-1k. This effect is conserved when linearly evaluating, suggesting that the effect is not simply a statistical coincidence. Finally, we observe that an increase in the scale of available pretraining data almost universally significantly improved transfer performance of RadNet-pretrained models, often even beyond ImageNet performance (cf. Figures 2 and 3, or Supplementary Tables ST1-3).

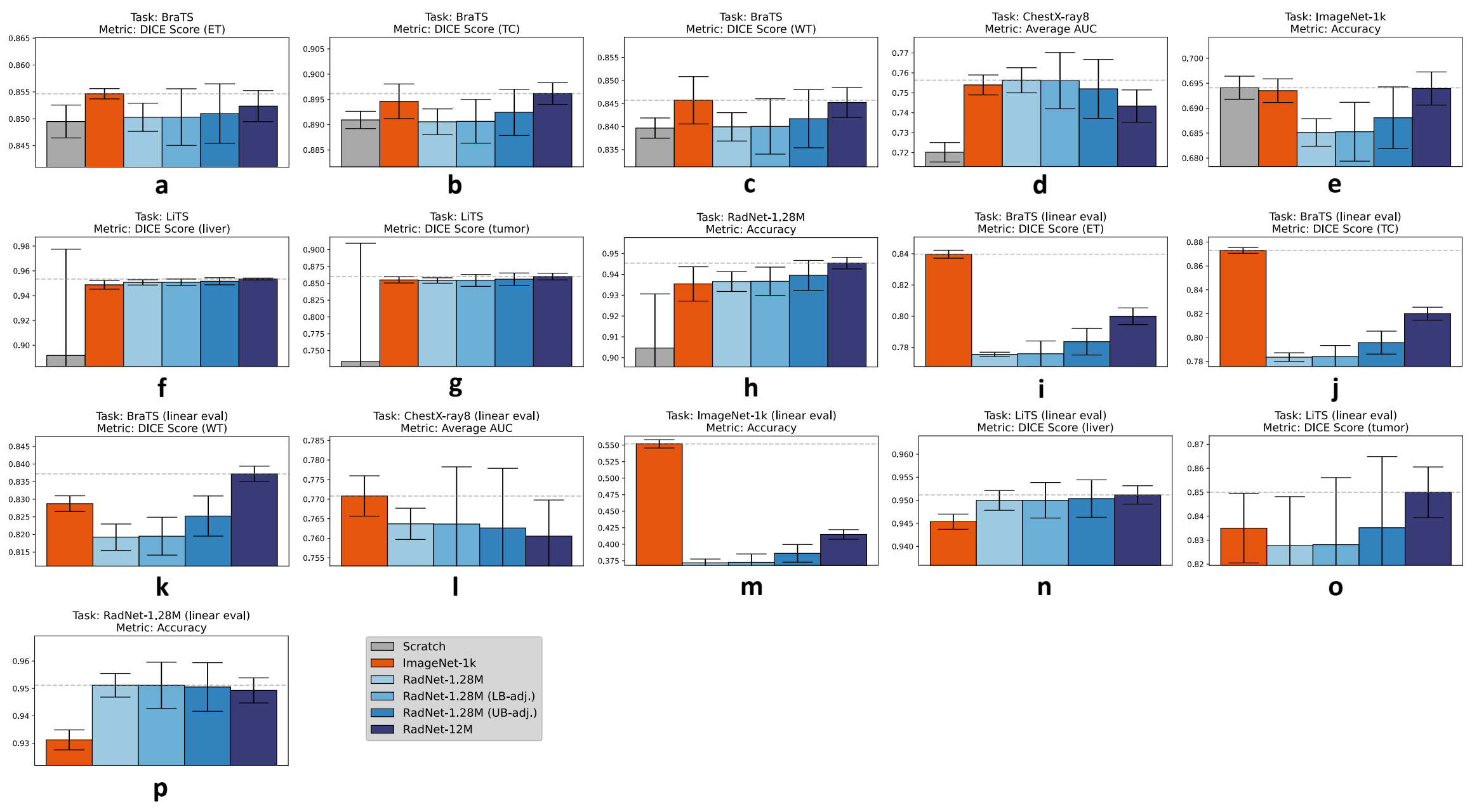

**Fig. 2: Intra-domain outperforms cross-domain transfer learning.**

Performance across four separate runs for every combination of pretraining dataset and fine-tuning task of Experiments #1 and #2 are on display: **a**, Fine-tuning on BraTS 2020 (Enhancing Tumor). **b**, Fine-tuning on BraTS 2020 (Tumor Core). **c**, Fine-tuning on BraTS 2020 (Whole Tumor). **d**, Fine-tuning on ChestX-Ray8. **e**, Fine-tuning on ImageNet-1k. **f**, Fine-tuning on LiTS 2017 (Liver). **g**, Fine-tuning on LiTS 2017 (Lesion). **h**, Fine-tuning on RadNet-1.28M. **i**, Linear evaluation on BraTS 2020 (Enhancing Tumor). **j**, Linear evaluation on BraTS 2020 (Tumor Core). **k**, Linear evaluation on BraTS 2020 (Whole Tumor). **l**, Linear evaluation on ChestX-Ray8. **m**, Linear evaluation on ImageNet-1k. **n**, Linear evaluation on LiTS 2017 (Liver). **o**, Linear evaluation on LiTS 2017 (Lesion). **p**, Linear evaluation on RadNet-1.28M.

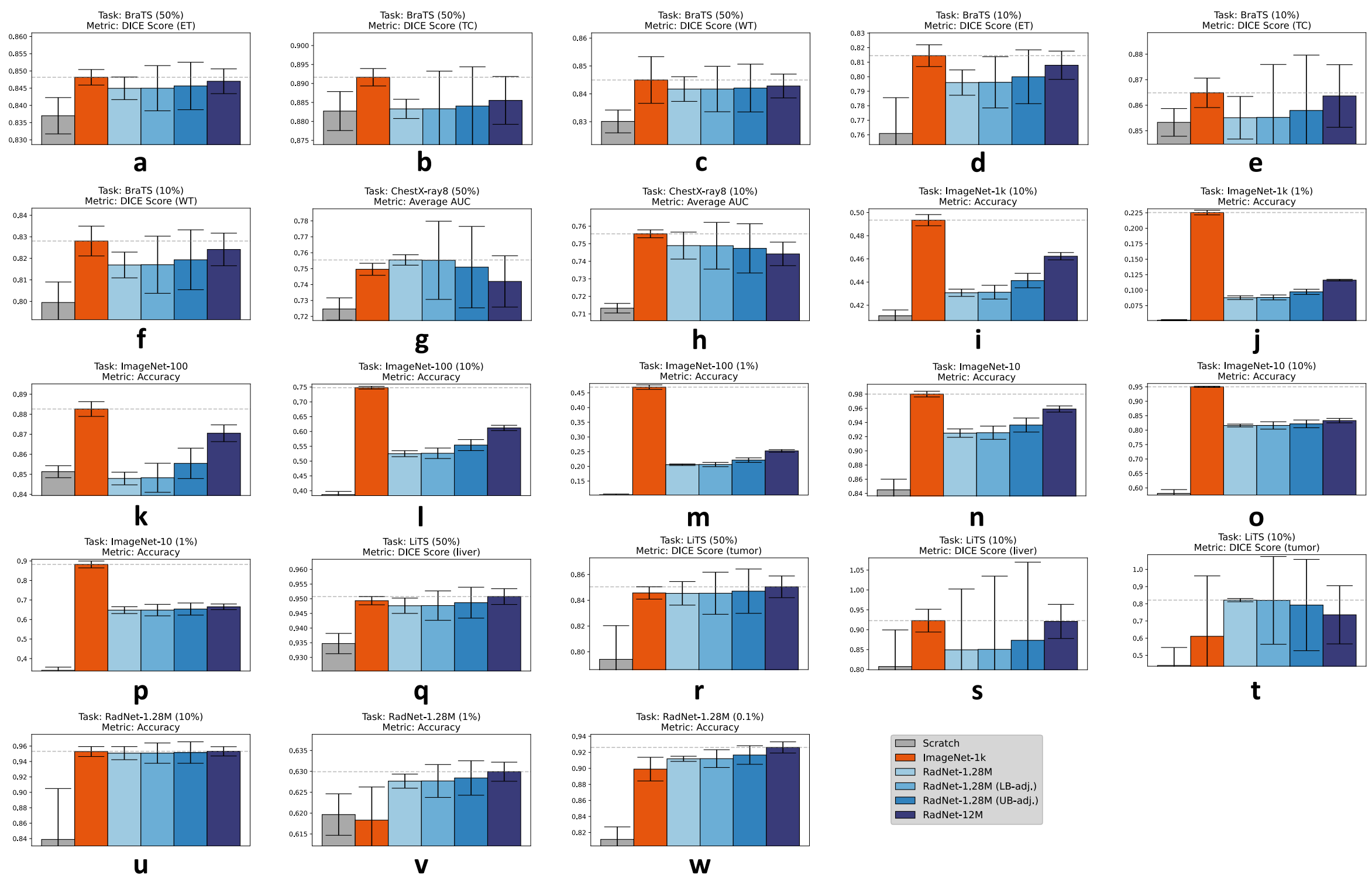

**Fig. 3: The intra-domain advantage scales with task complexity.**

Performance across four separate runs for every combination of pretraining dataset, fine-tuning task of Experiment #3 are on display: **a**, Fine-tuning on BraTS 2020 (50%, Enhancing Tumor). **b**, Fine-tuning on BraTS 2020 (50%, Tumor Core). **c**, Fine-tuning on BraTS 2020 (50%, Whole Tumor). **d**, Fine-tuning on BraTS 2020 (10%, Enhancing Tumor). **e**, Fine-tuning on BraTS 2020 (10%, Tumor Core). **f**, Fine-tuning on BraTS 2020 (10%, Whole Tumor). **g**, Fine-tuning on ChestX-Ray8 (50%). **h**, Fine-tuning on ChestX-Ray8 (10%). **i**, Fine-tuning on ImageNet-1k (10%). **j**, Fine-tuning on ImageNet-1k (1%). **k**, Fine-tuning on ImageNet-100. **l**, Fine-tuning on ImageNet-100 (10%). **m**, Fine-tuning on ImageNet-100 (1%). **n**, Fine-tuning on ImageNet-10. **o**, Fine-tuning on ImageNet-10 (10%). **p**, Fine-tuning on ImageNet-10 (1%). **q**, Fine-tuning on LiTS 2017 (50%, Liver). **r**, Fine-tuning on LiTS 2017 (50%, Lesion). **s**, Fine-tuning on LiTS 2017 (10%, Liver). **t**, Fine-tuning on LiTS 2017 (10%, Lesion). **u**, Fine-tuning on RadNet-1.28M (10%). **v**, Fine-tuning on RadNet-1.28M (1%). **w**, Fine-tuning on RadNet-1.28M (0.1%).

# 4 Discussion

From our results, we conclude that cross-domain transfer indeed degrades performance compared to intra-domain transfer learning in accordance with Hypothesis I, particularly in few-shot or linear evaluation scenarios. For CT-based tasks, this intra-domain advantage is relatively small, compared to natural image-based tasks. In order to observe it with high statistical significance, multiple runs of the same experiment can be necessary. However, despite similar test-time performance, the train-time loss falls at markedly different speeds (cf. Fig. 4) between intra-domain, cross-domain and no pretraining. This suggests that the proposed intra-domain advantage, while probably extant, is suppressed. This explanation is also corroborated by the fact that the expected intra-domain advantage for both data domains is visible and amplified substantially for most few-shot fine-tuning and linear evaluation scenarios.

The asymmetrical generalization gap we observe follows an intuitive logic: information encoded in the color channels has different meanings between RadNet and ImageNet-pretraining. Whereas models pretrained on ImageNet can reuse any texture-encoding feature and "discard" color-encoding features in a cross-domain transfer scenario, RadNet-pretrained models must learn an entirely new concept in the reverse case. The overall competitive performance of ImageNet, and the asymmetry in generalization gaps which we observe, substantiate our second Hypothesis; ImageNet pretraining delivers a set of features that on average transfer much better than RadNet features, though they do not generally outperform intra-domain transfer.

Another implication of our results is that data domains possess very narrow definitions. Judging by their performance, RadNet's CT images are part of the same subdomain as the LiTS CT images, but not as the BraTS MRI images. If there exists such a thing as a medical image domain, its subdomains (CTs, MRIs, X-rays, etc.) are distinct entities divided by generalization gaps. As CTs and MRIs rely on different physics and display the same anatomical structure somewhat differently, this distinction is quite intuitive. Consequentially, if the choice is between a capable generalist model and a specialist medical model from a different domain, the generalist model, i.e. an ImageNet-pretrained model, appears to be the better choice.

Curiously, the convergence speed in terms of training loss and validation performance both appear to correlate only partially with the pretraining domain, sometimes offering a similar and sometimes offering a slightly improved convergence speed. Generally, any pretraining appears to increase convergence speed on any task. However, more performant models occasionally require more time to converge or decrease their training loss at a slower or similar rate, compared to less performant ones (for example during fine-tuning on ImageNet-1k). The rate of training loss decrease and convergence speed measured by validation performance also appear to be only loosely correlated to each other. This implies that eventual test-time performance should not be estimated by observing either of these quantities.

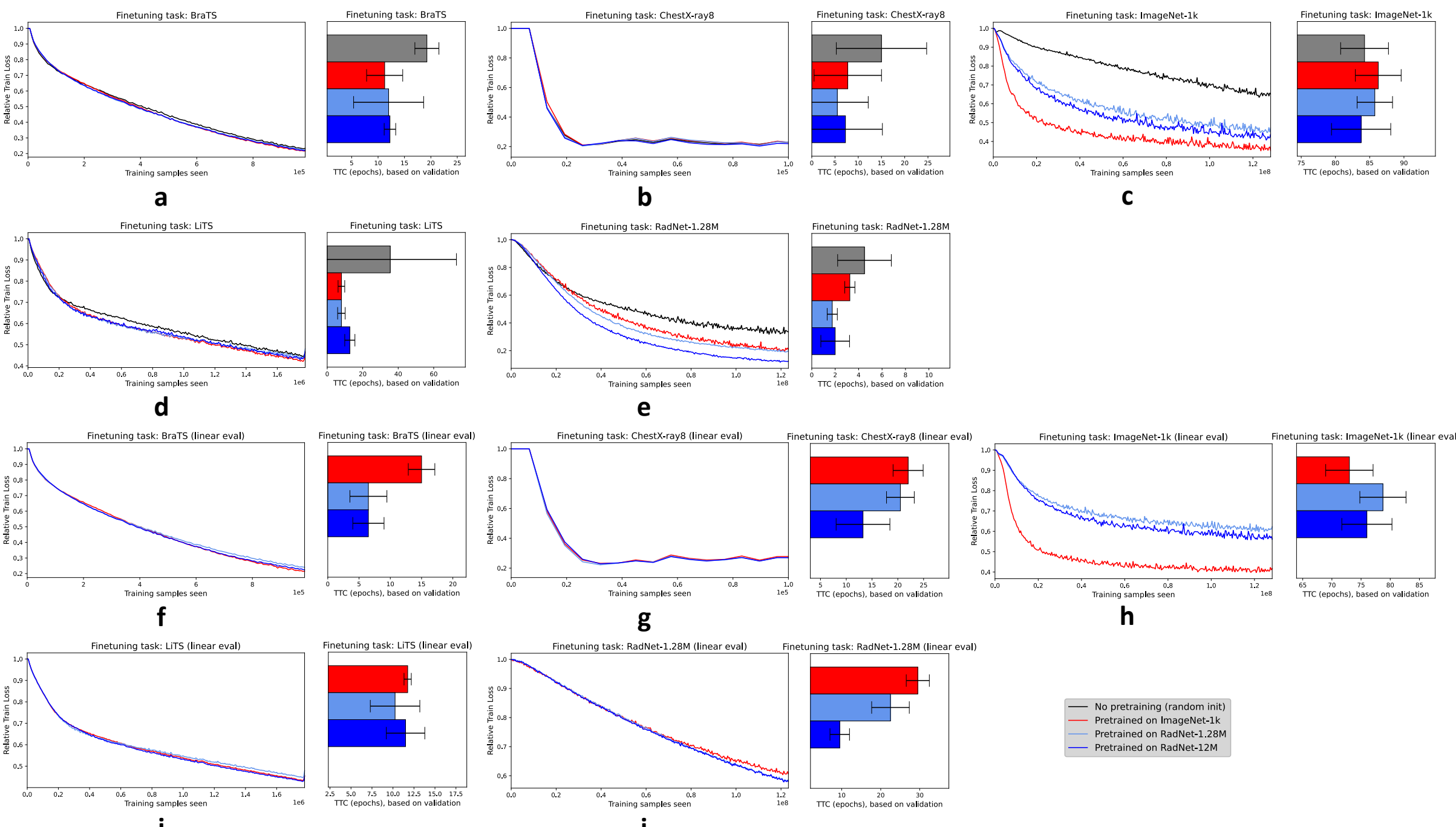

**Fig. 4: Models generally converge more quickly after intra-domain transfer.**
Convergence speed is estimated in two ways for each combination of pretraining dataset and task. Firstly, via training loss decrease (left graph), and secondly via time-to-convergence (TTC), which is approximated as the first epoch in which validation performance has not improved by at least 0.5% for a patience duration of 10 epochs, analogously to common early stopping methods. The figure displays these estimates for fine-tuning on BraTS 2020 (**a**), ChestX-ray8 (**b**), ImageNet-1k (**c**), LiTS 2017 (**d**) and RadNet-1.28M (**e**). It also displays these estimates for linear evaluation on BraTS 2020 (**f**), ChestX-ray8 (**g**), ImageNet-1k (**h**), LiTS 2017 (**i**) and RadNet-1.28M (**j**). Performance improvement still sometimes occurs after this point, although only with barely visible changes in validation loss. A pattern of inverse correlation between performance and convergence speed estimates is observed, although not across all experiments.

## 4.1 Additional feature space analysis

Fig. 5 depicts the degree of feature reuse during transfer learning, approximating the similarity of lowest-level convolution filters before and after fine-tuning using Centered Kernel Alignment [19]. We find that feature reuse is greater in intra-domain than cross-domain transfer scenarios for the ImageNet (significantly) and RadNet (slightly) classification tasks. Like our observations on performance, reduced feature reuse reaffirms that CT and MRI scans constitute separate data domains. Additionally, we observe a higher average feature reuse across tasks for ImageNet pretraining compared to RadNet pretraining further corroborating Hypothesis II. Interestingly, this feature reuse remains high even when the color channels no longer encode color (CTs) or encode different information (MRIs). This implies that for early convolution kernels, ImageNet pretraining encodes shape and texture more strongly than color, which is in agreement with previous observations in literature [20].

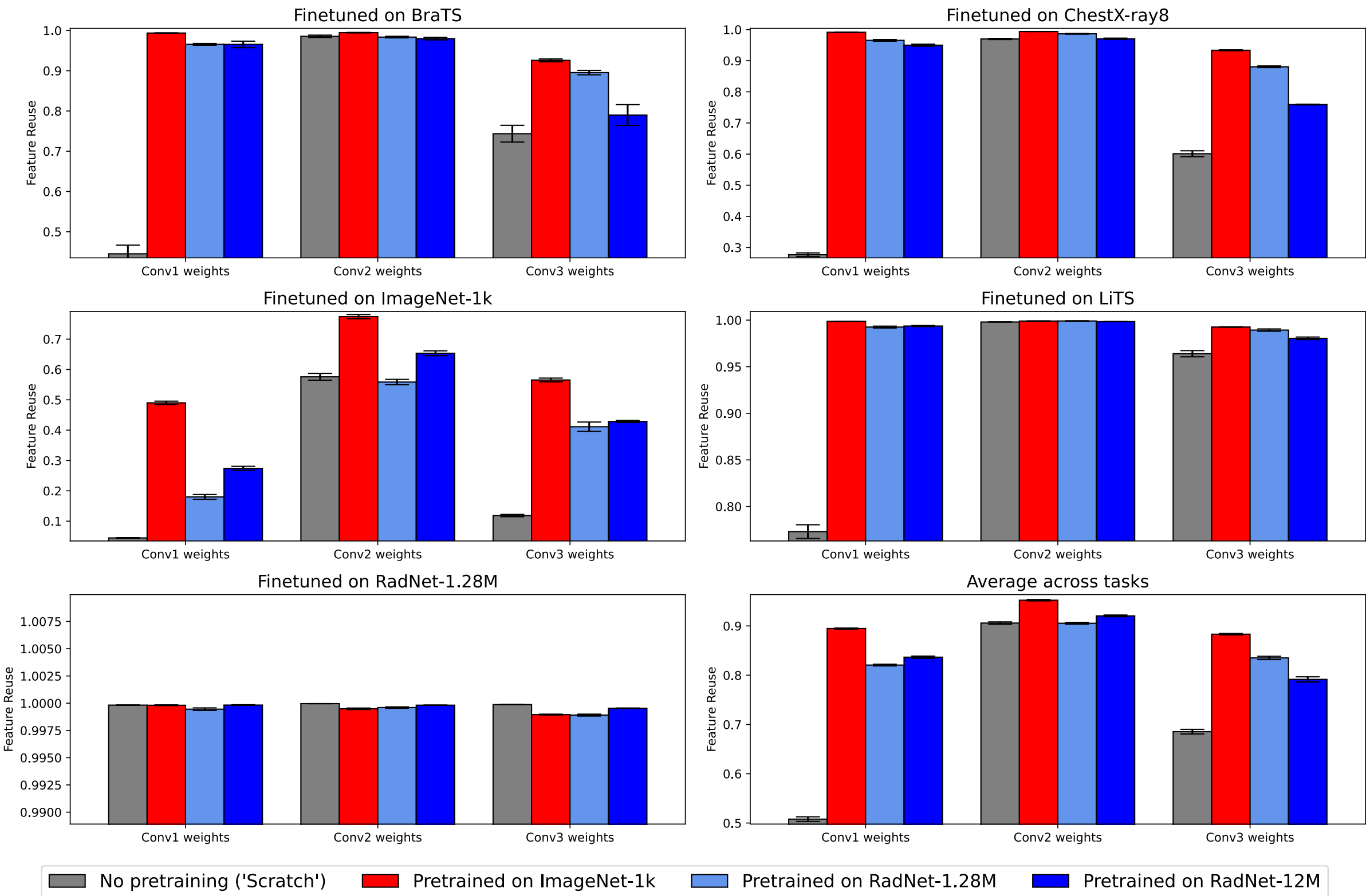

**Fig. 5: Intra-domain transfer leads to increased feature reuse.**
Increased feature reuse is evident for intra-domain transfer scenarios. Feature reuse is measured for the first three convolution layers of the pretrained and fine-tuned models (using RBF-kernel CKA). It is visibly increased in intra-domain transfer scenarios, although feature reuse after RadNet pretraining is only marginally higher than that of ImageNet pretraining when fine-tuning on RadNet-1.28M. Notably, even when channels encode different information, such as when pretraining happens on ImageNet and fine-tuning on BraTS, this feature reuse remains high – higher than for pretraining on CT images.

## 4.2 The nature of the intra-domain transfer performance gains

To understand where the performance gains of RadNet pretraining are found in practice, we investigated two possibilities: I) Unique low-level features which models acquire during in-domain pretraining are responsible for the intra-domain advantage. In this case we expect to see many data points which only the model with the intra-domain advantage correctly identifies and few or no data points which only the other models correctly identify. II) Alternatively, the classification process could be so intractable that such features cannot be isolated. In that case, whether a data point is correctly identified should depend only on a model's known performance but not the data point itself. Fig. 6 displays histograms of the fraction of data points correctly identified by the combination of models described on the X-axis. For example, the bar labeled "ImageNet" shows the number of data points identified correctly only by the ImageNet-pretrained model, but not by any other models. The corresponding hatched bars show the expected numbers, assuming the second explanation is correct.

We observe that two models with similar performance do not necessarily correctly classify the same data. Similarly, models with superior performance do not correctly classify a strict superset of data compared to models of inferior performance. This observation is in agreement with previous observations by Cohen et al. [8]. While there exists for every model at least a small number of data points which only that model could correctly identify, this number is always highest for the model(s) pretrained on the task domain (rightmost bars in Figs. 6a-d). This favors our first explanation that unique features are responsible for the intra-domain advantage. This effect is again substantially magnified during linear evaluation.

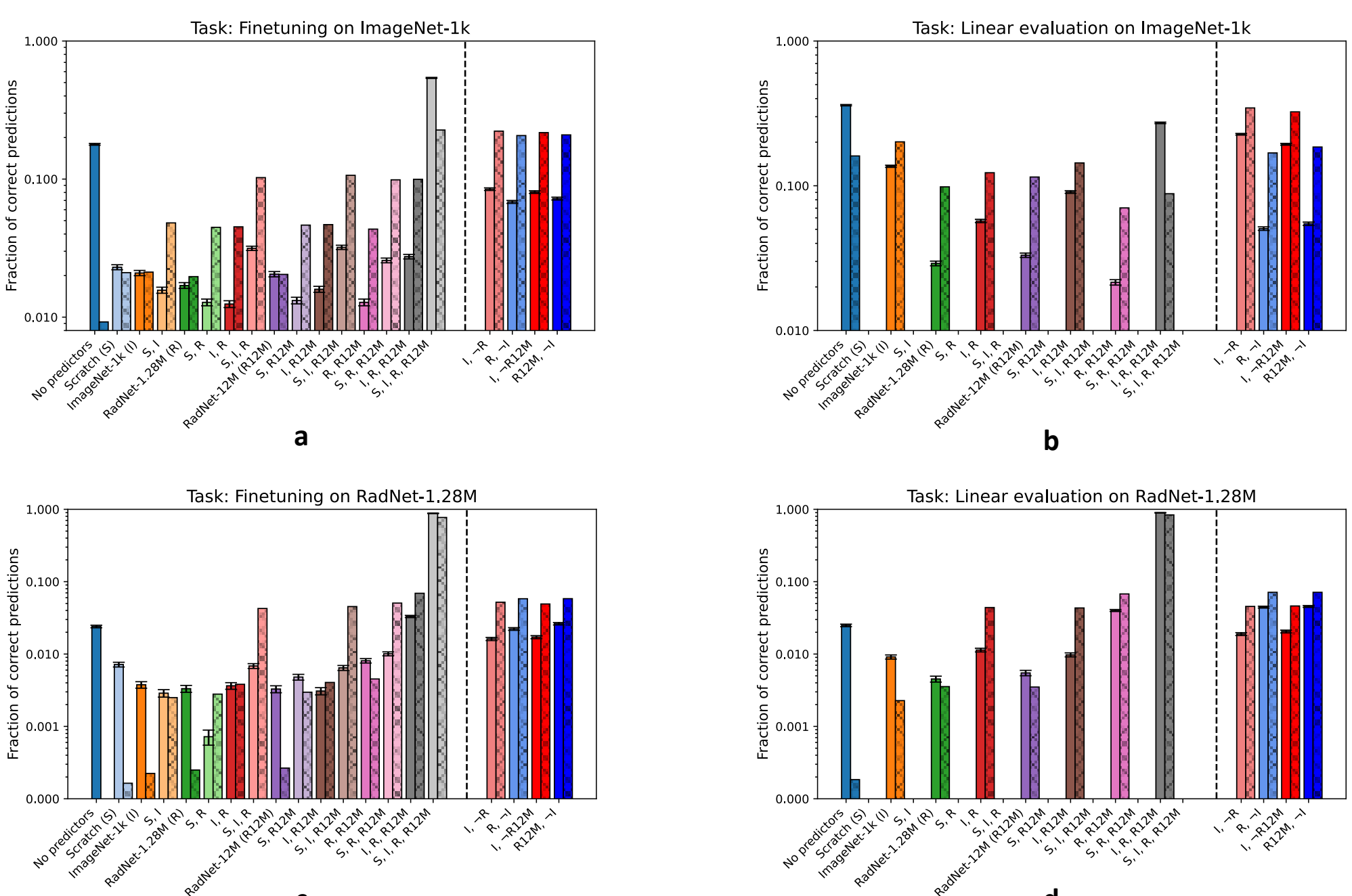

**Fig. 6: Intra-domain transfer leads to development of unique features and meaningful feature reuse.** This figure encodes the fraction of the fine-tuning dataset was correctly predicted by which pretraining sources (i.e. the bar labeled “RadNet, ¬ImageNet” shows the fraction of the test set identified correctly only by the RadNet-pretrained model, but not (¬) by the ImageNet-pretrained model). This is shown for: **a**, Fine-tuning on ImageNet-1k. **b**, Linear evaluation on ImageNet-1k. **c**, Fine-tuning on RadNet-1.28M. **d**, Linear evaluation on RadNet-1.28M. An intra-domain advantage exists, with a significant number of data points only predicted by the model pretrained on data of the same domain. This effect is amplified in the linear evaluation scenario.

Through investigation of around 60 LiTS images with high prediction discrepancies between RadNet- and ImageNet-pretrained models, we found that an appreciable part of the performance gains for the LiTS liver segmentation corresponded to explainable, macroscopic image/region properties. Improvements from RadNet pretraining are often observed in border regions between classes and for (semi-)insular segmentation targets like seemingly unconnected pieces of liver tissue, whose connection to the liver is vertical. These conditions seem intuitively reasonable: insular areas of organ tissue are likely to appear in the training set during RadNet pretraining and less likely during ImageNet pretraining. Furthermore, contrast-enhancing features must be learned during RadNet pretraining to identify structures, whereas ImageNet-pretraining can do the same using only texture or color information. This potentially gives RadNet-pretrained models an advantage near region borders on the colorless CT images of LiTS or RadNet, where the unenhanced contrast is low. Fig. 7 shows several representative examples of this advantage.

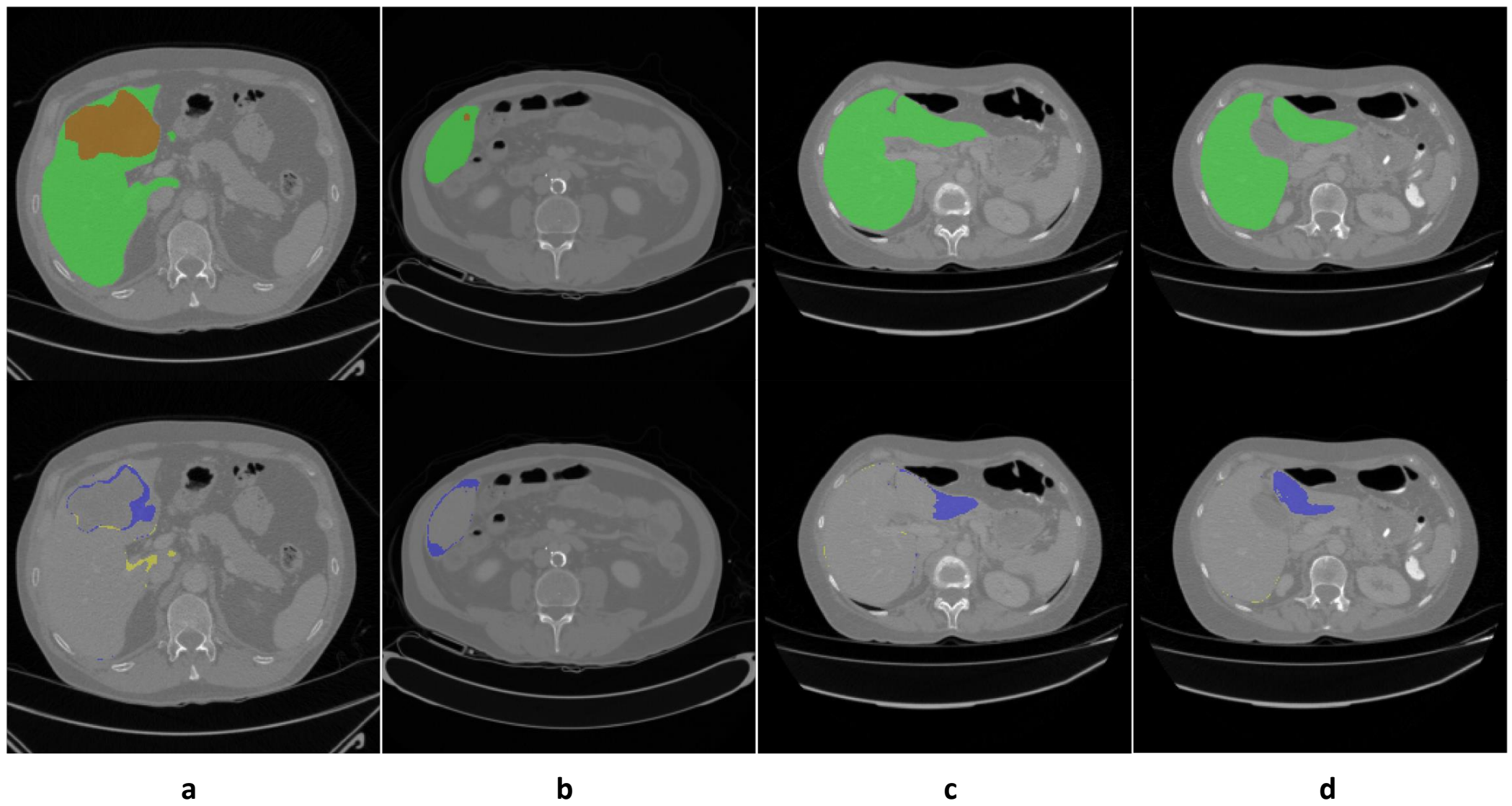

**Fig. 7: RadNet's intra-domain advantage relates to explainable, macroscopic image properties.**
Four example images from the LiTS 2017 segmentation task are shown, with colored areas highlighting pixels correctly classified only by the RadNet-pretrained model, but not the ImageNet-pretrained model (blue) and the reverse case (yellow). Each image corresponds to one or two of the three typical scenarios in which RadNet pretraining appears to outperform ImageNet pretraining: **a**, Border region between liver and tumor. **b**, Border region between liver and surrounding tissue. **c**, Semi-insular region of the liver. **d**, Insular region of the liver. Note that the images shown are some of the more extreme examples - most of the time, the differences in predictions are much more minute.

## 5 Conclusions & Outlook

In this paper, we have systematically explored the efficacy of self-supervised pretraining and transfer learning across multiple datasets from the natural and medical image domains. Our experiments demonstrate the existence of a performance-degrading generalization gap that occurs during cross-domain transfer learning. They further highlight that different combinations of pretraining and fine-tuning datasets yield differently sized gaps, depending on task domain, complexity, and whether or not linear evaluation is performed. We confirmed our initial hypotheses that I) intra-domain pretraining is generally preferable, even though II) some datasets such as ImageNet generalize better on average. We have provided evidence that in-domain pretraining yields unique low-level features which are preserved during fine-tuning on data from the same domain, and have discussed the nature of the performance gains they provided. Finally, we have made available our pretrained and fine-tuned models as baselines for future work.

Several limitations apply to this work. Firstly, our conclusions may not generalize to different hyperparameters. Furthermore, models trained using other pretraining algorithms (e.g. generative or generative-adversarial methods), supervision levels, or data domains different from ours may not support our conclusions, although there is no obvious reason why our a priori assumptions would not apply. The same limitation extends to other, yet to be tested model architectures. Modern Transformer [21] or State Space models [22], for example, may exhibit different behaviors. It should also be noted that our results for linear evaluation on segmentation tasks should be viewed with some degree of caution, as the addition of the trainable U-Net decoder still allowed half the model to be trained, potentially offsetting some intra-domain advantage and introducing a source of uncertainty. Finally, all data in the RadNet datasets comes from one hospital. This limitation may have somewhat reduced the generalizability of the dataset, potentially implying a larger intra-domain advantage than we report in this paper.

We summarize a general set of rules for optimal transfer learning performance, based on our experiments:

- Intra-domain transfer learning typically performs similarly or outperforms cross-domain transfer learning by at least a small margin. In short, "**My medical AI should be looking at pictures of tumors instead of birds to learn about tumors.**"
- Intra-domain pretraining is generally preferable in few-shot or linear evaluation settings.
- The definition of a data domain can be very narrow. Our experiments suggest that intra-domain transfer for MR images would require different pretraining data compared to CT images, for example.
- We observed no performance saturation for further upscaling of the pretraining in many scenarios, with RadNet-12M-pretrained models often outperforming ImageNet-1k-pretrained models. Thus, **a wealth of previously unused, unlabeled medical images can in fact be leveraged to achieve scaling performance gains**.
- We observed that the generalization gap varies very nuancedly; Even the target class or metric, e.g. liver vs. lesion on the LiTS task, can be relevant for the choice of pretraining dataset.
- Using ImageNet as a default pretraining dataset is suitable. ImageNet pretraining gave models remarkable training stability and competitive performance across most tasks in our experiments. Hence, we recommend its use if in-domain pretraining data is not available in sufficient quantities, and always at least testing its performance.
- **There is no "one size fits all" solution.** If the best possible performance on a task is more relevant than a quick solution, there is no substitute for testing multiple pretraining datasets, particularly from the same data domain as the task.

In future work, we plan to extend RadNet-12M with additional data from other modalities and to pretrain additional architectures using RadNet. We believe that research extending our work to other pretraining algorithms and supervised scenarios would constitute a promising research direction, addressing some of the limitations of this work. Creating large-scale models pretrained on a previously untapped wealth of unlabeled clinical images would be highly valuable to the exponentially growing, yet notoriously data-starved [23] field of medical computer vision.

## Code availability

The pretraining and fine-tuning code framework developed for this paper, including configuration files, code used to create figures, and results logs can be found at https://github.com/TIO-IKIM/Transfer-learning-across-domain-boundaries. Our code, and some publicly available code we build on, are subject to the MIT license. For publicly available datasets, we link the datasets and provide our data layout/splits.

## Data availability

We link the sources of the public datasets we used in our code repository. The RadNet datasets are not made public. All pretrained/fine-tuned models reported are linked on https://github.com/TIO-IKIM/Transfer-learning-across-domain-boundaries and may be used for further research. This permission includes models pretrained on the RadNet datasets.

## Acknowledgements

We acknowledge KITE (Plattform für KI-Translation Essen) from the REACT-EU initiative (EFRE-0801977, https://kite.ikim.nrw/).

# Supplementary Materials

## S1 - Significance of experimental results

We report the significance of performance gains or losses based on Welch's t-test [18]. This test quantifies the probability that two populations of samples with different mean and standard deviation were sampled from the same underlying distribution. In layman's terms, the resulting p-value is the probability that our performance gain comes from a real effect in the experiment and is not random noise. Figs. SF1-3 contain the relative performance gains/losses between any pairing of pretraining options on a specific fine-tuning task and their corresponding p-value. Note that the p-values for the LB/UB-adjusted RadNet results generally have increased uncertainties compared to the unadjusted results. This additional uncertainty stems from the interpolation process and error propagation and does not imply any sort of training instability for RadNet-pretrained models.

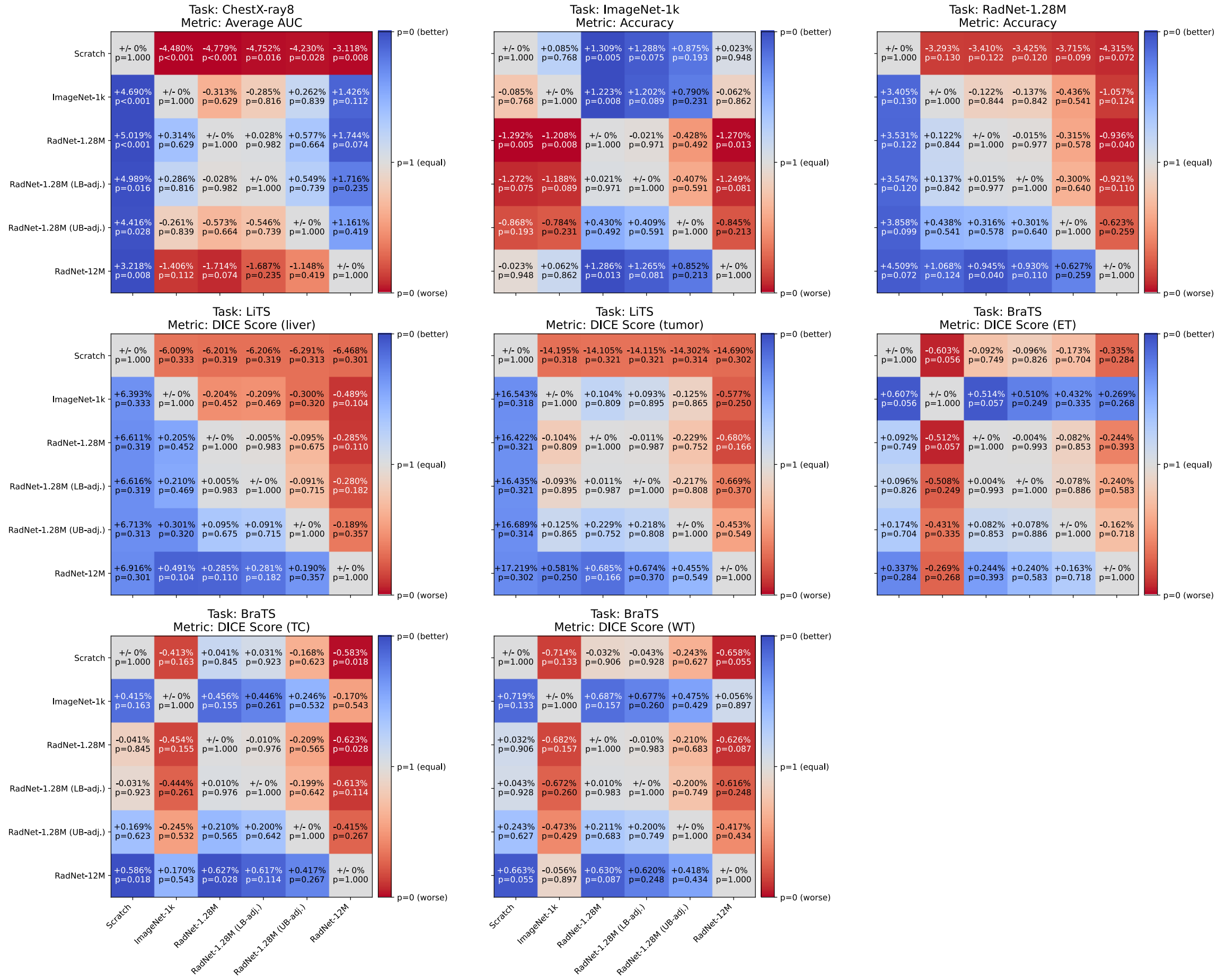

**Fig. SF1: Relative performance and significance of results in Experiment 1.**
In each subplot, each possible pair of pretraining options is compared once for the task and performance metric specific to that subplot. The p-values are derived from Welch's t-test. To read a subplot, apply the following logic: for the given task and metric, pretraining on the dataset in this row is better/worse than pretraining on the dataset in this column by some percentage and with a significance of the given p-value. Note that the t-test is two-tailed and thus shows the probability that the observed differences are random chance, not that one pretraining method is better than the other. For the latter, a one-tailed t-test would be appropriate, yielding probabilities that are half as high, and thus twice as significant.

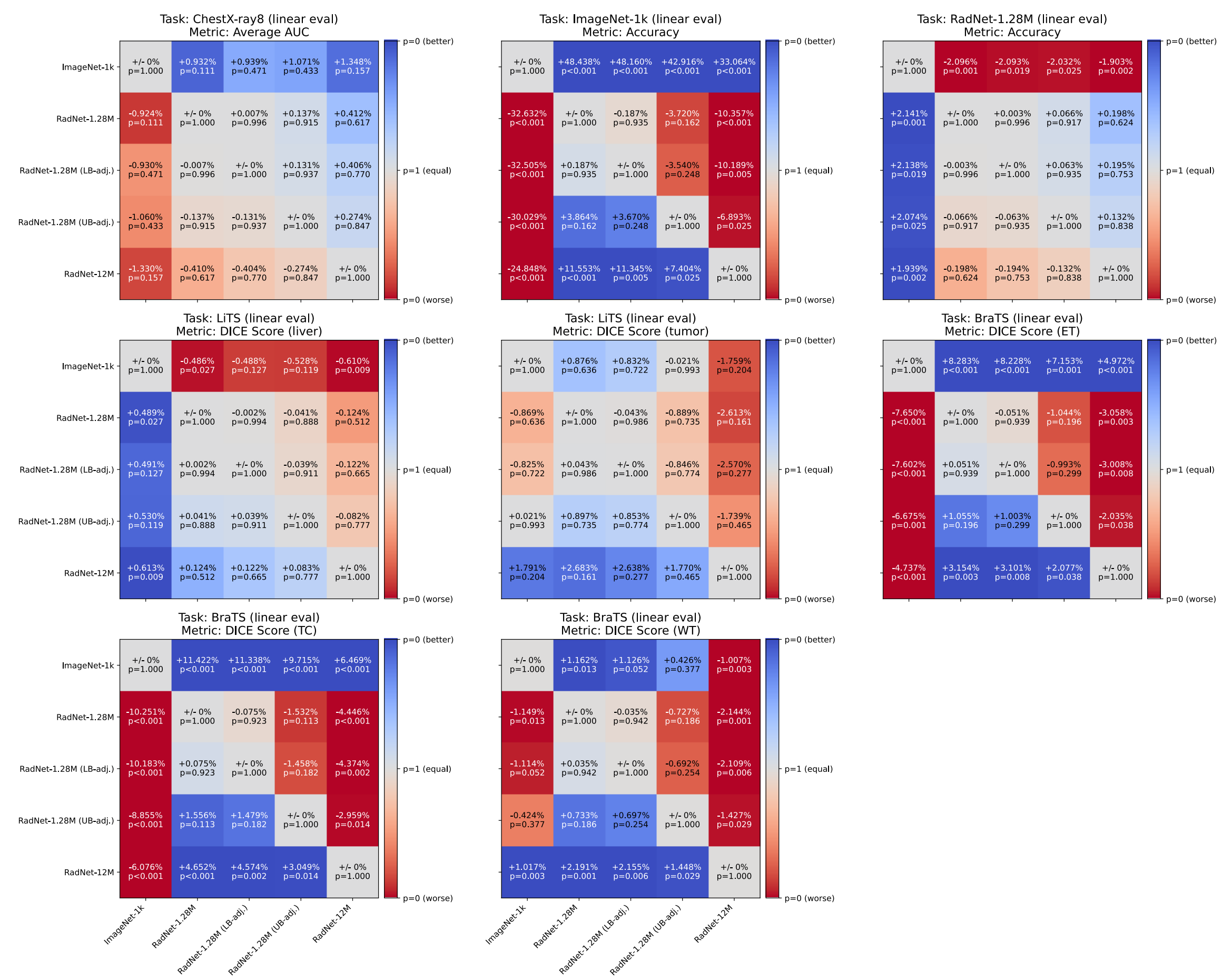

**Fig. SF2: Relative performance and significance of results in Experiment 2.**

In each subplot, each possible pair of pretraining options is compared once for the task and performance metric specific to that subplot. The p-values are derived from Welch's t-test. To read a subplot, apply the following logic: for the given task and metric, pretraining on the dataset in this row is better/worse than pretraining on the dataset in this column by some percentage and with a significance of the given p-value. Note that the t-test is two-tailed and thus shows the probability that the observed differences are random chance, not that one pretraining method is better than the other. For the latter, a one-tailed t-test would be appropriate, yielding probabilities that are half as high, and thus twice as significant.

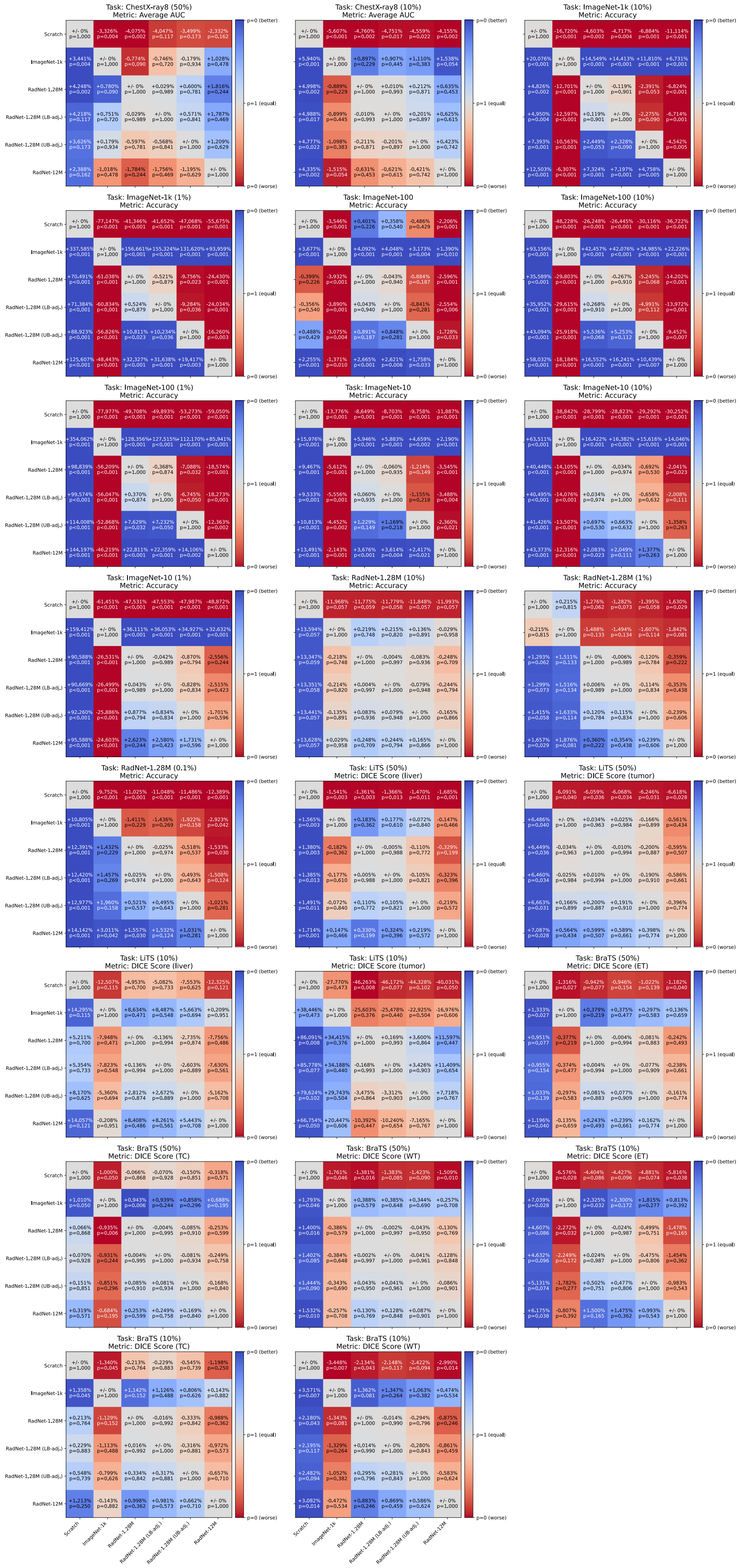

**Fig. SF3: Relative performance and significance of results in Experiment 3.**

In each subplot, each possible pair of pretraining options is compared once for the task and performance metric specific to that subplot. The p-values are derived from Welch's t-test. To read a subplot, apply the following logic: for the given task and metric, pretraining on the dataset in this row is better/worse than pretraining on the dataset in this column by some percentage and with a significance of the given p-value. Note that the t-test is two-tailed and thus shows the probability that the observed differences are random chance, not that one pretraining method is better than the other. For the latter, a one-tailed t-test would be appropriate, yielding probabilities that are half as high, and thus twice as significant.

## S2 - Experimental setup

To pretrain, we apply SimCLR [24] on a ResNet-50 [25] architecture. Self-supervised pretraining is chosen to circumvent the annotation bottleneck of medical data. A contrastive approach is chosen over generative/adversarial approaches due to the sensitivity of generative methods to rare samples [26], and the propensity for training instability of adversarial methods [26]. SimCLR is chosen over other contrastive approaches for its widespread adoption and low model overhead. However, since our initial assumptions about data distributions of different imaging domains are independent of the pretraining approach, we expect our results to generalize to other pretraining methods as well.

The ResNets are randomly initialized. Pretraining is performed with a batch size of 4096, using the LARS optimizer [27]. For the learning rate, we apply the square root rule used by Chen et al. [24], yielding a learning rate of 4.8. All input images are resized to 3 x 256 x 256 pixels so that all models possess the same number of parameters, facilitating fair comparison between the datasets. For RadNet images this is achieved by concatenating the same slice three time along the channel axis. All images are normalized to the range [0, 1]. All models are pretrained on their respective datasets for 100 epochs, using NT-CrossEntropy Loss [24]. To prevent pretraining instability from information leakage, our ResNet-50 uses SyncBatchNorm [28] in place of BatchNorm [29].

For the fine-tuning, the SimCLR MLP head is replaced by a new head. To use the pretrained models, RadNet, LiTS 2017 and ChestX-Ray8 images are converted to three-channel images as described above. For BraTS 2020, the T2 channel is discarded for the same reason (discarding the T2-channel empirically had the smallest performance impact). Fine-tuning is performed with a common set of hyperparameters per individual fine-tuning task, regardless of the pretraining dataset, to allow for a fair comparison (for a detailed discussion on this matter, refer to the supplementary materials). Where necessary for segmentation, we extend the ResNets into a symmetrical U-Net-like [30] architecture. Newly added parts of a model are randomly initialized with PyTorch [31] default settings. All loss functions used are CrossEntropyLoss or a variation thereof (pixel-wise cross entropy for segmentations, multi-label cross entropy for ChestX-Ray8).

## S3 - On hyperparameter choice

One can make a fair comparison between two pretraining datasets in two ways. Either one argues that both competing training procedures are maximally optimized (which cannot be guaranteed without an exhaustive hyperparameter search) and therefore represent the best outcome the training procedure can offer. In this case, hyperparameters for networks pretrained on different data may be different on the same downstream task, if those parameters really do squeeze out that extra .1 percent of accuracy.

Alternatively, one can argue that both competing training procedures are not optimized, using arbitrary hyperparameters which converge with a reasonable performance when training from scratch and which are constant for any given downstream task. However, this may favor one of the competitors (which one specifically is unclear). Whether it actually does is impossible to know without a sufficiently fine-grained mapping of the entire hyperparameter space.

Given the prohibitive computational cost of such endeavors, we opt not to perform any large hyperparameter searches, and instead to make an arbitrary (but sufficiently stable and performant) choice of hyperparameters for each downstream task's fine-tuning process. We create as fair a comparison as possible, by first testing each task with a few learning rates and decay factors on an untrained model, and then reusing that parameter set.

Performance metrics reported in this paper may fall short of literature equivalents where such a hyperparameter search or extensive auxiliary procedures (additional regularization, auxiliary heads, optimizer choice, complex learning rate scheduling, etc.) were performed by the authors (although they remain surprisingly competitive on some tasks, considering this handicap). We abstain from using pre-existing hyperparameter heuristics or auxiliary procedures where possible, as those were typically developed by observing effectiveness in a specific domain (usually ImageNet). One exception to this approach is the use of the square root learning rate heuristic for LARS and on ImageNet, as optimization using LARS was necessary to reduce the computation time requirements to within reasonable time scales. The hyperparameters for all runs can be found in Tab. ST1 or at https://github.com/TIO-IKIM/Transfer-learning-across-domain-boundaries

It is important to note that while the comparisons presented in our experiments are fair in the sense of unbiased sampling of the hyperparameter space, they are just that: a sample in a very high-dimensional space. It is therefore possible that the behavior we report is different in other areas of this parameter space and that the conclusions we draw do not generalize, although preliminary experiments with other parameters do not suggest that this is the case. Similarly, our results depend on the available data, so that a varied dataset with more medical imaging modalities or more data may produce different results. The scale of our experiments implies that neither is the case, but does offers no guarantee.

## S4 - Correlation between neighboring slices in RadNet

Neighboring slices of one CT image are necessarily similar, since (thankfully) organs cannot have random gaps or slices offset in random directions. Hence, already having neighboring slices in the dataset reduces the new information a slice added to the dataset can provide during training. To some degree, this is unavoidable because human anatomy is a strongly constraining factor. However, even compared to a dataset in which every slice comes from the CT scan of a different, random patient, some information is redundant. While we cannot provide a dataset consisting completely of uncorrelated slices, we can scale up our initial dataset until the information content of the new dataset is the same as that of an entirely uncorrelated dataset the size of our initial dataset, so that only domain-specific redundancy remains.

We will try to motivate an approximate lower and upper bound for a scale factor $S$, by which the amount of data in RadNet-1.28M should be multiplied to achieve a fairer comparison. In our results, we report the performances of a hypothetical RadNet-1.28M scaled with the lower and upper bound for $S$. These performances are estimated using interpolation.

To arrive at an approximate lower bound $S_{LB}$, we test the compressibility of a set of correlated images (i.e. an entire 3D CT scan) compared to an equally-sized set of random images drawn from RadNet. Any gain in compressibility is most likely due to the exploitation of correlated image features, implying a reduced information content in the correlated images. This calculation yields a lower bound because the compression algorithm is not guaranteed to be able to exploit every correlation - in practice, $S$ is likely to be much larger. To measure compressibility, we used Python's gzip compression package [32] (at default settings). We calculate the compressibility ratio across all images (weighted by the number of slices) in RadNet-1.28M and find: $S_{LB} = \sum_{i=1}^{90663} w_i \frac{C_{corr.}(i)}{C_{random}} \approx 1.136$ .

To arrive at an upper bound $S_{UB}$, we utilize the concept of mutual information (MI) [33]. $MI$ quantifies the amount of information about a variable $X_1$ which can be inferred by observing a correlated variable $X_2$. Note that $MI$ does not specifically capture representational similarity in the way a human observer might understand, only a mathematical measure of "shared information" (unlike CKA which was expressly designed for the former purpose).

For every combination of an image and a random reference image, we can establish a relative $MI$ between two slices of the former image as follows:

- Calculate the $MI$ between every combination of two image slices of the first image (In theory, this calculation can be performed for all combinations, but beyond around 10 slices, the $MI$ will already be quite small)
- Normalize these $MI$ values by using the $MI$ between a slice and itself as $MI_{max}$ and the average $MI$ between the image slice and slices from the reference image as $MI_{min}$. The resulting relative utility value $U$, with $U_{i,j} = 1 - \frac{MI_{i,j} - MI_{min}}{MI_{max} - MI_{min}}$, is a measure of the usefulness of a slice, compared to a random one. The closer $U_{i,j}$ is to 0, the less new information that slice will give us. A utility of 0 indicates that a slice with an identical distribution (in a 2d-histogram) already exists in our data, while a 1 indicates that the new slice is as new as a slice from a random reference image.

We can then calculate our upper bound $S_{UB}$ as the sum of the worst utility $U_{i,max}$ for every slice i, divided by the number of slices in our dataset. In order not to "count double", slices are added to the dataset one by one, calculating the worst utility with respect to the dataset at the time that the new image is added.

Our calculation yields an upper bound because the average $MI$ between slices of the same CT image is likely, albeit not guaranteed, to stem from anatomy-related correlation (since we have already subtracted the average $MI$ that we expect to exist between two slices of different CT images, which should cover aspects all CT images have in common, like the bed or air around the patient). For RadNet-1.28M, we find that $S_{UB} \approx 3.807$.

A typical distribution of $U$ vs. frame distance (for all slice pairings in a single image and a reference image) is shown in Fig. SF4. Note that this distribution varies from image to image depending on reference image, captured body region, filters, and frame distance.

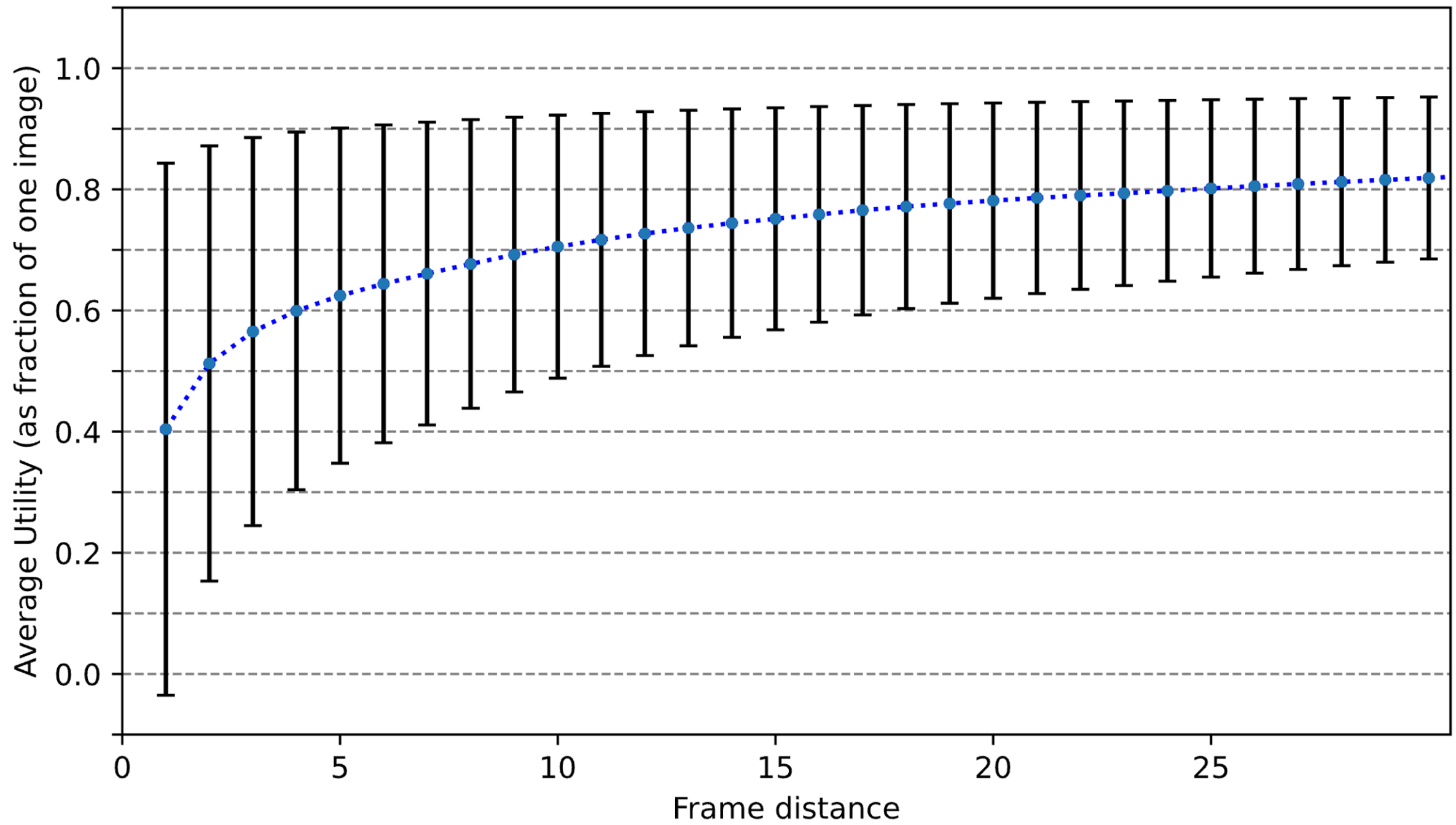

**Fig. SF4: Typical distribution of image utility in a single 3D CT scan.**
In a 3D CT scan, the closer an image slice is to its neighbor, the more information it contains about its neighbor. As seen in this figure, the actual value of an image is significantly decreased if its direct neighbor is already in the dataset. The curve shown was computed for a CT with a frame distance of 5mm. Its shape is representative for most scans, although scan region, frame distance, and image quality cause significant variations.

Note that the upper bound represents a more realistic approximation of $S$: The underlying 2d-histograms of the $MI$ calculation are very finely binned (64 bins wide along both axes vs. 4096 possible integer values each pixel can have in a typical CT), which implies that high $MI$ values do indeed represent highly similar images and not just vaguely similar histograms, while the lower bound is somewhat restricted in its potential height it can go because of things like noise (which is incompressible) and imperfect compression of existing similarities.

## S5 - Feature Uniqueness

By randomly splitting all features of a convolution layer into smaller segments and averaging over the CKA similarity [19] of all pairings, we can obtain a measure for the uniqueness of the features obtained during pretraining (cf. Fig. SF5). Surprisingly, we find that RadNet-based pretraining yields substantially greater feature uniqueness than ImageNet-based pretraining, in spite of working without any color information. A possible explanation for this phenomenon is two-fold. Firstly, given a number of (as established above) correlated images, a part of the features created will likely be the result of overfitting to the differences between two or more highly correlated images. These features, representative of statistical noise, will be highly unique, but not useful. This explanation is corroborated by the fact that the RadNet-1.28M-pretrained model has more unique features than the RadNet-12M-pretrained model. Secondly, as ImageNet pretraining encodes color information, the feature uniqueness of its respective model may be reduced if the model encodes the same shape, texture, or gradient multiple times in different color combinations, which it may have to do in some cases during pretraining. This would explain why ImageNet-1k encodes fewer unique features than both RadNet-pretrained models.

Between the increased feature uniqueness, and favorable performance in intra-domain transfer scenarios, we conclude that RadNet-based pretraining results in a number of highly specialized features, which increase performance on CT-based downstream tasks.

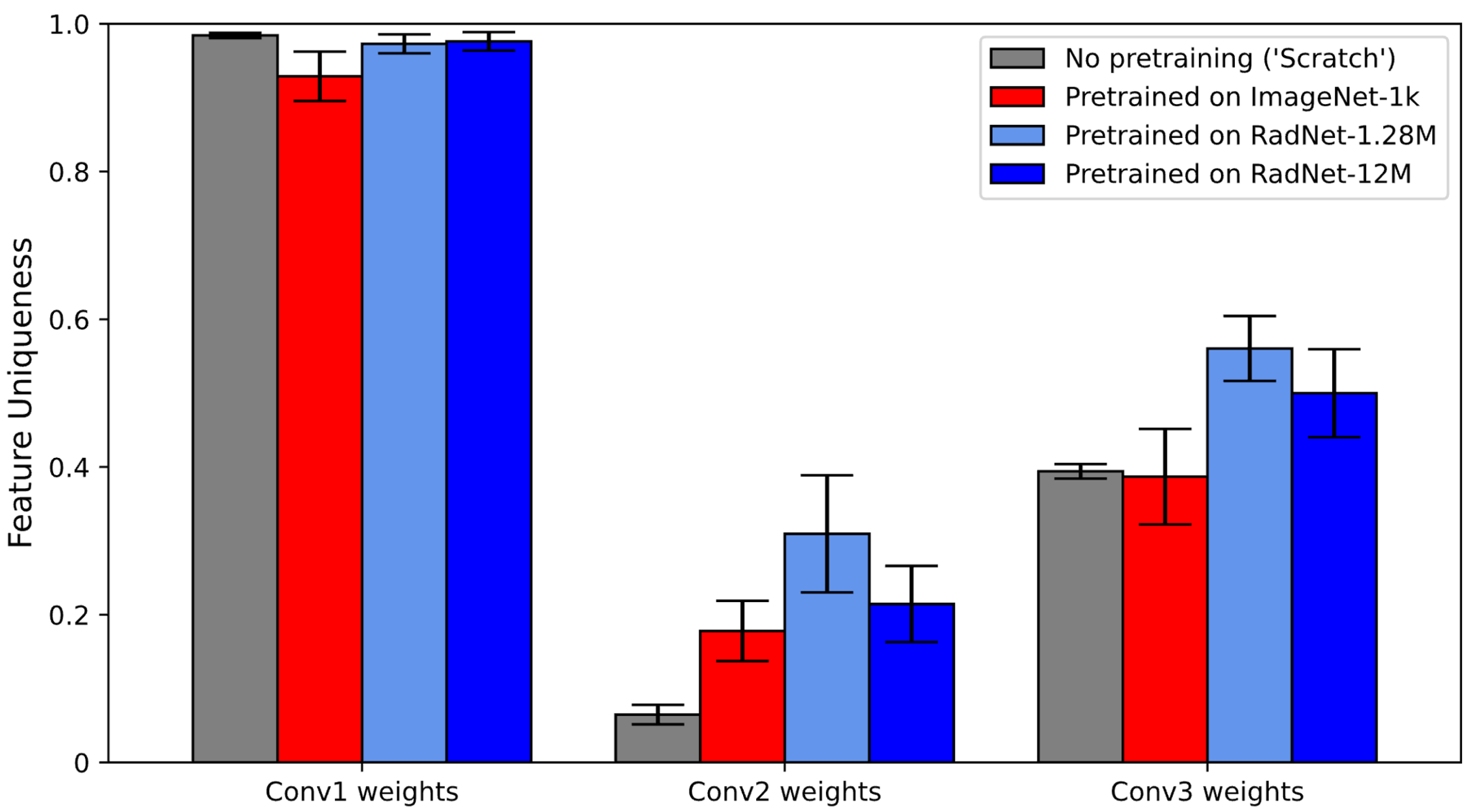

**Fig. SF5: Feature uniqueness of convolution layers after pretraining.**
For each model, feature uniqueness is quantified by comparing the similarity of the convolution kernels after pretraining with themselves, by way of splitting the set of all kernels of a convolution layer into subsets and comparing each individual pairing. The higher the uniqueness coefficient, the more the individual kernels differ from one another. Surprisingly, RadNet-pretrained models appear to have more unique features than the ImageNet-pretrained model.

## S6 – Mixed modality pretraining

One obvious direction of advance would be simply to combine multiple available pretraining modalities, following the rationale that the information within these datasets complements each other and creates one model that generalizes exceptionally well. It could then deliver the best performance on many or all modalities, out of the box. In this section, we repeat Experiment 1 on the RadNet and ImageNet downstream tasks, using mixed modality pretraining. To perform the mixed modality pretraining, we pretrain one model by pretraining on both datasets sequentially for 50 epochs each (ImageNet first, RadNet-1.28M second), and another on a 50/50 mixture of both datasets (stratified per dataset), for the full 100 epochs. We call these models Sequential (S) and Mixed (M). Interestingly, we observe that neither model appears to be able to beat the specialist models in either task (cf. Fig. SF6) despite the increased variety of training data. We observe that the Mixed model approaches or matches the performance of the specialist models, while the Sequential model generally underperforms the specialist models on either task – even in the domain in which it was last pretrained. As in the other experiments, when the difficulty is increased by performing linear evaluation instead of fine-tuning, we observe the same tendencies, but with much more pronounced gaps. This implies that the long duration of fine-tuning afforded during these experiments partially offsets the performance loss incurred during mixed modality pretraining, although the limited nature of this experiment does not permit definitive conclusions.

We posit that the main limitation at play here is the model's capacity to reflect the training data in the feature space. Essentially, the model is too small to learn from both data distributions without "overwriting" information. This idea is corroborated by the fact that the Mixed model outperforms the Sequential model on both tasks, since the Mixed model actively incentivizes not overwriting previous features, but rather finding features that generalize well. This generalizability is also reflected in the significantly increased feature reuse for the Mixed model compared to both the Sequential and the RadNet-pretrained specialist model.

As a general conclusion to this experiment, we recommend avoiding mixed modality pretraining when working with small models and downstream tasks whose modality is known ahead of time. Instead, it is preferable to opt for single modality pretrained specialist models, where they are available or feasible to train. Where mixed modality pretraining is desirable, this limited experiment suggests that finding a way to mix data from all source modalities is strongly preferable to sequential per-modality training of already pretrained models.

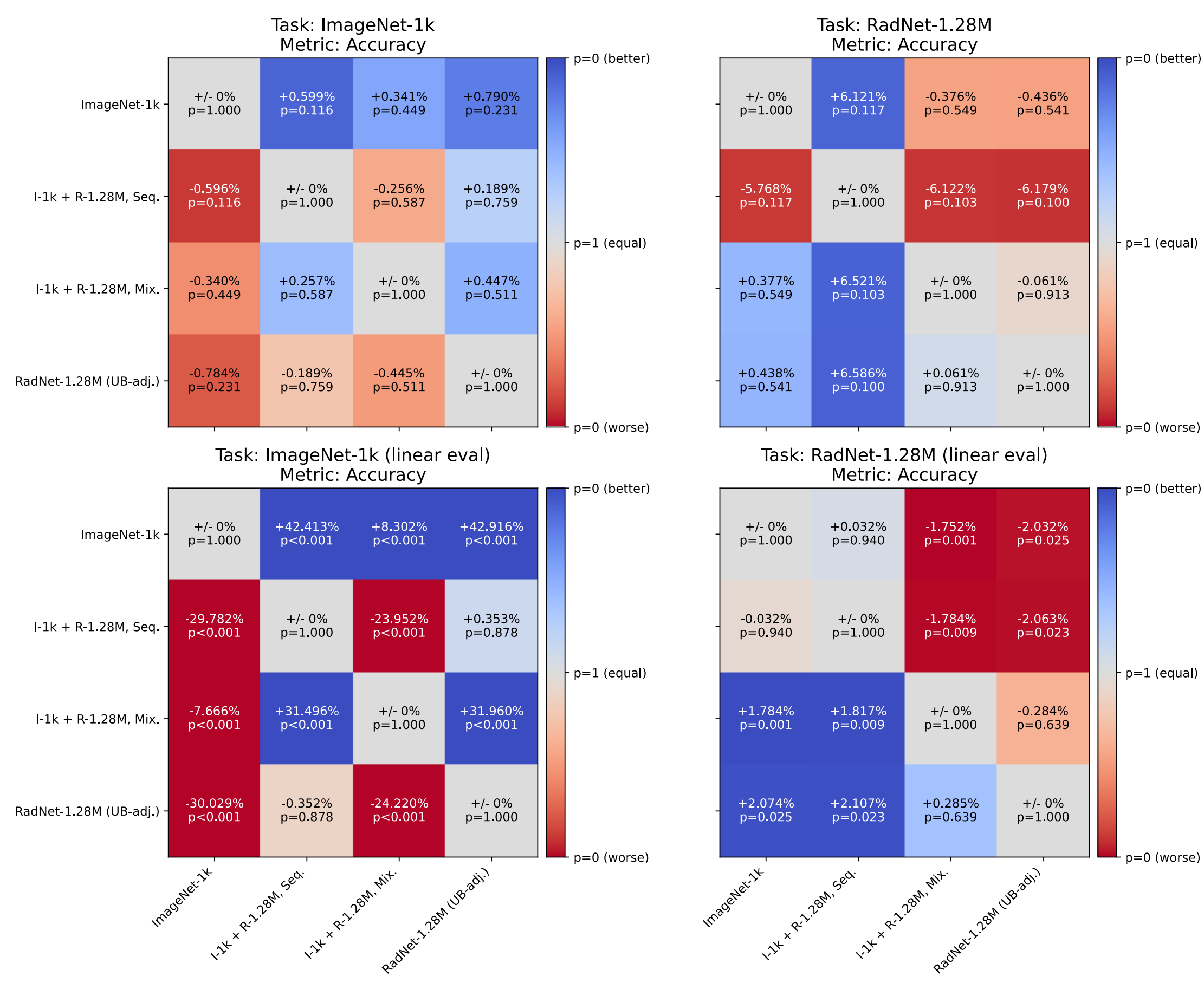

**Fig. SF6: Relative performance and significance of mixed modality pretraining results.**
In each subplot, each possible pair of pretraining options is compared once for the task and performance metric specific to that subplot. The p-values are derived from Welch's t-test. To read a subplot, apply the following logic: for the given task and metric, pretraining on the dataset in this row is better/worse than pretraining on the dataset in this column by some percentage and with a significance of the given p-value. Note that the t-test is two-tailed and thus shows the probability that the observed differences are random chance, not that one pretraining method is better than the other. For the latter, a one-tailed t-test would be appropriate, yielding probabilities that are half as high, and thus twice as significant.

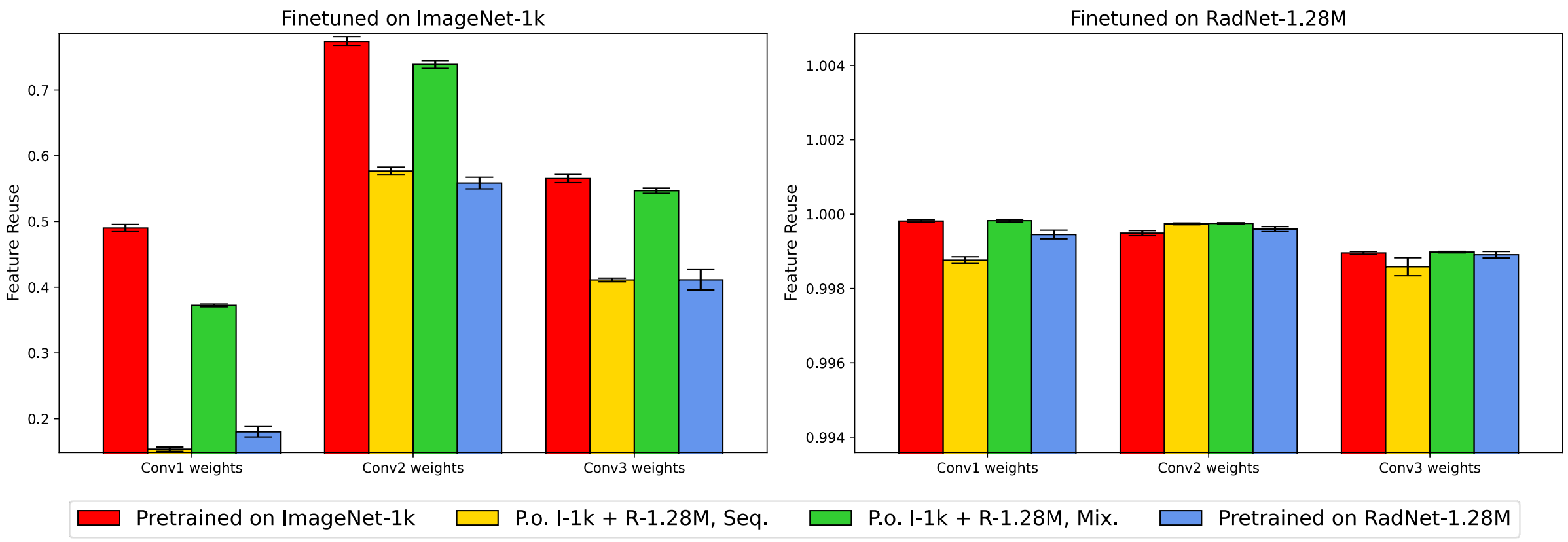

**Fig. SF7: Feature reuse after mixed modality pretraining strongly depends on the mixing.**
Feature reuse is measured for the first three convolution layers of the pretrained and fine-tuned models (using RBF-kernel CKA). "Seq" and "Mix" denote sequential and mixed pretraining, respectively, both using ImageNet-1k and RadNet-1.28M data. Notably, feature reuse is significantly higher when pretraining in a mixed fashion, compared to sequential pretraining. This remains true even on the RadNet-1.28M downstream task, although the second step of the sequential pretraining, which one might naively expect to "overwrite" learned features, occurs on RadNet-1.28M.

## Supplementary Tables

| Task (Metric) | No Pretraining | ImageNet-1k | RadNet-1.28M | RadNet-1.28M (LB-adj.) | RadNet-1.28M (UB-adj.) | RadNet-12M |
|---|---|---|---|---|---|---|
| BraTS 2020 (DICE, ET) | 0.8495 ± 0.0031 | **0.8547 ± 0.0010** | 0.8503 ± 0.0026 | 0.8503 ± 0.0053 | 0.8510 ± 0.0055 | 0.8524 ± 0.0029 |
| BraTS 2020 (DICE, TC) | 0.8909 ± 0.0017 | 0.8946 ± 0.0034 | 0.8906 ± 0.0026 | 0.8907 ± 0.0043 | 0.8924 ± 0.0046 | **0.8962 ± 0.0021** |
| BraTS 2020 (DICE, WT) | 0.8397 ± 0.0022 | **0.8457 ± 0.0051** | 0.8399 ± 0.0031 | 0.8400 ± 0.0060 | 0.8417 ± 0.0063 | 0.8452 ± 0.0033 |
| ChestX-ray8 (Avg. AUC) | 0.7202 ± 0.0049 | 0.7539 ± 0.0050 | **0.7563 ± 0.0063** | 0.7561 ± 0.0141 | 0.7520 ± 0.0148 | 0.7433 ± 0.0081 |
| ImageNet-1k (Accuracy) | **0.6941 ± 0.0023** | 0.6935 ± 0.0024 | 0.6851 ± 0.0028 | 0.6853 ± 0.0059 | 0.6881 ± 0.0062 | 0.6939 ± 0.0033 |
| LiTS 2017 (DICE, Liver) | 0.8917 ± 0.0857 | 0.9487 ± 0.0035 | 0.9507 ± 0.0021 | 0.9507 ± 0.0027 | 0.9516 ± 0.0028 | **0.9534 ± 0.0008** |
| LiTS 2017 (DICE, Tumor) | 0.7337 ± 0.1759 | 0.8551 ± 0.0046 | 0.8542 ± 0.0040 | 0.8543 ± 0.0088 | 0.8561 ± 0.0092 | **0.8600 ± 0.0050** |
| RadNet-1.28M (Accuracy) | 0.9046 ± 0.0260 | 0.9354 ± 0.0082 | 0.9365 ± 0.0048 | 0.9367 ± 0.0068 | 0.9395 ± 0.0072 | **0.9454 ± 0.0028** |

**Table ST1: Experiment 1 Results.**
Results for Experiment 1. All results are averages from four runs with different random initializations. Best pretraining options are marked in bold. Relative performance increases/decreases for all pairings and their statistical significances can also be found in the supplementary materials (S1).

| Task (Metric) | No Pretraining | ImageNet-1k | RadNet-1.28M | RadNet-1.28M (LB-adj.) | RadNet-1.28M (UB-adj.) | RadNet-12M |
|---|---|---|---|---|---|---|
| BraTS 2020 linear eval. (DICE, ET) | N/A | **0.8397 ± 0.0026** | 0.7755 ± 0.0014 | 0.7759 ± 0.0082 | 0.7837 ± 0.0086 | 0.7999 ± 0.0054 |
| BraTS 2020 linear eval. (DICE, TC) | N/A | **0.8730 ± 0.0024** | 0.7835 ± 0.0037 | 0.7841 ± 0.0092 | 0.7957 ± 0.0096 | 0.8200 ± 0.0054 |
| BraTS 2020 linear eval. (DICE, WT) | N/A | 0.8287 ± 0.0022 | 0.8192 ± 0.0037 | 0.8195 ± 0.0054 | 0.8252 ± 0.0057 | **0.8372 ± 0.0022** |
| ChestX-ray8 linear eval. (Avg. AUC) | N/A | **0.7708 ± 0.0052** | 0.7637 ± 0.0040 | 0.7636 ± 0.0146 | 0.7626 ± 0.0152 | 0.7606 ± 0.0092 |
| ImageNet-1k linear eval. (Accuracy) | N/A | **0.5520 ± 0.0063** | 0.3718 ± 0.0058 | 0.3725 ± 0.0128 | 0.3862 ± 0.0134 | 0.4148 ± 0.0073 |
| LiTS 2017 linear eval. (DICE, Liver) | N/A | 0.9454 ± 0.0017 | 0.9500 ± 0.0021 | 0.9500 ± 0.0039 | 0.9504 ± 0.0041 | **0.9512 ± 0.0020** |
| LiTS 2017 linear eval. (DICE, Tumor) | N/A | 0.8350 ± 0.0145 | 0.8278 ± 0.0204 | 0.8281 ± 0.0280 | 0.8352 ± 0.0297 | **0.8500 ± 0.0106** |
| RadNet-1.28M linear eval. (Accuracy) | N/A | 0.9312 ± 0.0036 | **0.9511 ± 0.0043** | 0.9511 ± 0.0084 | 0.9505 ± 0.0089 | 0.9493 ± 0.0046 |

**Table ST2: Experiment 2 Results.**
Results for Experiment 2. All results are averages from four runs with different random initializations. Best pretraining options are marked in bold. Relative performance increases/decreases for all pairings and their statistical significances can also be found in the supplementary materials (S1).

| **Task (Metric)** | No Pretraining | ImageNet-1k | RadNet-1.28M | RadNet-1.28M (LB-adj.) | RadNet-1.28M (UB-adj.) | RadNet-12M |
|---|---|---|---|---|---|---|
| BraTS 2020 50% data (DICE, ET) | 0.8370 ± 0.0053 | 0.8482 ± 0.0023 | 0.8450 ± 0.0033 | 0.8450 ± 0.0066 | 0.8456 ± 0.0069 | **0.8470 ± 0.0036** |
| BraTS 2020 50% data (DICE, TC) | 0.8827 ± 0.0051 | **0.8916 ± 0.0023** | 0.8833 ± 0.0025 | 0.8833 ± 0.0099 | 0.8841 ± 0.0103 | 0.8855 ± 0.0063 |
| BraTS 2020 50% data (DICE, WT) | 0.8301 ± 0.0041 | **0.8450 ± 0.0084** | 0.8417 ± 0.0044 | 0.8417 ± 0.0082 | 0.8421 ± 0.0086 | 0.8428 ± 0.0043 |
| BraTS 2020 10% data (DICE, ET) | 0.7609 ± 0.0247 | **0.8145 ± 0.0075** | 0.7960 ± 0.0087 | 0.7962 ± 0.0176 | 0.8000 ± 0.0185 | 0.8079 ± 0.0097 |
| BraTS 2020 10% data (DICE, TC) | 0.8533 ± 0.0054 | **0.8649 ± 0.0058** | 0.8551 ± 0.0083 | 0.8552 ± 0.0207 | 0.8579 ± 0.0217 | 0.8636 ± 0.0123 |
| BraTS 2020 10% data (DICE, WT) | 0.7995 ± 0.0095 | **0.8280 ± 0.0069** | 0.8169 ± 0.0060 | 0.8170 ± 0.0133 | 0.8193 ± 0.0139 | 0.8241 ± 0.0076 |
| ChestX-ray8 50% data (Avg. AUC) | 0.7247 ± 0.0070 | 0.7496 ± 0.0038 | **0.7555 ± 0.0033** | 0.7552 ± 0.0246 | 0.7510 ± 0.0256 | 0.7420 ± 0.0161 |
| ChestX-ray8 10% data (Avg. AUC) | 0.7133 ± 0.0027 | **0.7557 ± 0.0023** | 0.7489 ± 0.0077 | 0.7489 ± 0.0133 | 0.7474 ± 0.0140 | 0.7442 ± 0.0067 |
| ImageNet-1k 10% data (Accuracy) | 0.4109 ± 0.0050 | **0.4934 ± 0.0048** | 0.4308 ± 0.0031 | 0.4313 ± 0.0059 | 0.4413 ± 0.0062 | 0.4623 ± 0.0032 |
| ImageNet-1k 1% data (Accuracy) | 0.0515 ± 0.0009 | **0.2254 ± 0.0038** | 0.0878 ± 0.0030 | 0.0883 ± 0.0040 | 0.0973 ± 0.0042 | 0.1162 ± 0.0014 |
| ImageNet-100 100% data (Accuracy) | 0.8513 ± 0.0030 | **0.8826 ± 0.0037** | 0.8479 ± 0.0032 | 0.8483 ± 0.0073 | 0.8555 ± 0.0076 | 0.8705 ± 0.0042 |
| ImageNet-100 10% data (Accuracy) | 0.3872 ± 0.0104 | **0.7479 ± 0.0046** | 0.5250 ± 0.0101 | 0.5264 ± 0.0177 | 0.5541 ± 0.0186 | 0.6119 ± 0.0090 |
| ImageNet-100 1% data (Accuracy) | 0.1034 ± 0.0025 | **0.4695 ± 0.0076** | 0.2056 ± 0.0026 | 0.2064 ± 0.0073 | 0.2213 ± 0.0076 | 0.2525 ± 0.0044 |
| ImageNet-10 100% data (Accuracy) | 0.8450 ± 0.0151 | **0.9800 ± 0.0040** | 0.9250 ± 0.0059 | 0.9256 ± 0.0094 | 0.9364 ± 0.0099 | 0.9590 ± 0.0044 |
| ImageNet-10 10% data (Accuracy) | 0.5810 ± 0.0131 | **0.9500 ± 0.0020** | 0.8160 ± 0.0049 | 0.8163 ± 0.0128 | 0.8217 ± 0.0134 | 0.8330 ± 0.0077 |
| ImageNet-10 1% data (Accuracy) | 0.3400 ± 0.0162 | **0.8820 ± 0.0175** | 0.6480 ± 0.0174 | 0.6483 ± 0.0294 | 0.6537 ± 0.0310 | 0.6650 ± 0.0145 |
| LiTS 2017 50% data (DICE, Liver) | 0.9347 ± 0.0035 | 0.9493 ± 0.0014 | 0.9476 ± 0.0026 | 0.9477 ± 0.0050 | 0.9487 ± 0.0053 | **0.9507 ± 0.0027** |
| LiTS 2017 10% data (DICE, Liver) | 0.8074 ± 0.0921 | **0.9228 ± 0.0287** | 0.8495 ± 0.1530 | 0.8507 ± 0.1842 | 0.8734 ± 0.1966 | 0.9209 ± 0.0429 |
| LiTS 2017 50% data (DICE, Tumor) | 0.7942 ± 0.0261 | 0.8457 ± 0.0048 | 0.8454 ± 0.0091 | 0.8455 ± 0.0163 | 0.8471 ± 0.0172 | **0.8504 ± 0.0084** |
| LiTS 2017 10% data (DICE, Tumor) | 0.4413 ± 0.1045 | 0.6110 ± 0.3513 | **0.8212 ± 0.0094** | 0.8198 ± 0.2552 | 0.7927 ± 0.2655 | 0.7359 ± 0.1692 |
| RadNet-1.28M 10% data (Accuracy) | 0.8389 ± 0.0660 | 0.9529 ± 0.0064 | 0.9508 ± 0.0085 | 0.9509 ± 0.0132 | 0.9516 ± 0.0139 | **0.9532 ± 0.0060** |
| RadNet-1.28M 1% data (Accuracy) | 0.6197 ± 0.0050 | 0.6183 ± 0.0079 | 0.6277 ± 0.0017 | 0.6277 ± 0.0039 | 0.6284 ± 0.0041 | **0.6299 ± 0.0023** |
| RadNet-1.28M 0.1% data (Accuracy) | 0.8114 ± 0.0156 | 0.8990 ± 0.0148 | 0.9119 ± 0.0032 | 0.9121 ± 0.0111 | 0.9167 ± 0.0116 | **0.9261 ± 0.0070** |

**Table ST3: Experiment 3 Results.**

Results for Experiment 3. All results are averages from four runs with different random initializations. Best pretraining options are marked in bold. Relative performance increases/decreases for all pairings and their statistical significances can also be found in the supplementary materials (S1). Note that the sharp decrease in accuracy on the RadNet-1.28M (1%) task, compared to both more and less available data is likely an effect of the random choice of fine-tuning data. Interestingly, while the accuracy is worse than that of 0.1% data fine-tuning, the intra-domain advantage is still observed as normal, with the relative performance increase slightly below that of the 0.1% data scenario, as expected. Note that particularly in the low data regime, training instability causes generally higher variances in this experiment. This happens due to intentionally not using advanced augmentation strategies (typically developed on and for ImageNet), as those may offer domain-specific advantages that we do not wish to factor into our observations.

| Dataset name (pretraining) | Batch size | Learning rate | Optimizer | Method | # of images |
|---|---|---|---|---|---|
| ImageNet-1k | 4096 | 4.8 | LARS | SimCLR | 1,281,167 |
| RadNet-1.28M | 4096 | 4.8 | LARS | SimCLR | 1,281,167 |
| RadNet-12M | 4096 | 4.8 | LARS | SimCLR | 12,034,617 |
| **Task name (Experiments 1, 2)** | **Batch size** | **Learning rate** | **Optimizer** | **Decay factor** | **Class weights** |
| BraTS 2020 | 128 | 1.0e-3 | AdamW [34] | 0.95 | Background: 1<br>Others: 5 |
| ChestX-Ray8 | 128 | 5.0e-4 | AdamW | 0.97 | BG: 1e-3<br>Others: 1 |
| ImageNet-1k | 4096 | 4.8 | LARS | 0.97 | Equal |
| LiTS 2017 | 128 | 2.0e-4 | AdamW | 0.97 | BG: 1<br>Liver: 3<br>Lesion: 10 |
| RadNet-1.28M | 4096 | 1.0e-2[+] | LARS | 0.97 | Equal |
| **Task name (Experiment 3)** | **Batch size** | **Learning rate** | **Optimizer** | **Decay factor** | **Class weights** |
| BraTS 2020 (50% data) | 128 | 1.0e-3 | AdamW | 0.95 | BG: 1<br>Others: 5 |
| BraTS 2020 (10% data) | 128 | 1.0e-3 | AdamW | 0.95 | BG: 1<br>Others: 5 |
| ChestX-Ray8 (50% data) | 128 | 5.0e-4 | AdamW | 0.97 | BG: 1e-3<br>Others: 1 |
| ChestX-Ray8 (10% data) | 128 | 5.0e-4 | AdamW | 0.97 | BG: 1e-3<br>Others: 1 |
| ImageNet-1k (10% data) | 1024 | 2.4 | LARS | 0.97 | Equal |
| ImageNet-1k (1% data) | 256 | 1.0e-4 | AdamW | 0.97 | Equal |
| ImageNet-100 | 1024 | 2.4 | LARS | 0.97 | Equal |

| ImageNet-100 (10% data) | 256 | 1.0e-4 | AdamW | 0.97 | Equal |
|---|---|---|---|---|---|
| ImageNet-100 (1% data) | 256 | 1.0e-4 | AdamW | 0.97 | Equal |
| ImageNet-10 | 256 | 1.0e-4 | AdamW | 0.97 | Equal |
| ImageNet-10 (10% data) | 256 | 1.0e-4 | AdamW | 0.97 | Equal |
| ImageNet-10 (1% data) | 64 | 1.0e-4 | AdamW | 0.97 | Equal |
| LiTS 2017 (50% data) | 128 | 2.0e-4 | AdamW | 0.97 | BG: 1<br>Liver: 3<br>Lesion: 10 |
| LiTS 2017 (50% data) | 128 | 2.0e-4 | AdamW | 0.97 | BG: 1<br>Liver: 3<br>Lesion: 10 |
| RadNet-1.28M (10% data) | 1024 | 5.0e-3 | LARS | 0.97 | Equal |
| RadNet-1.28M (1% data) | 256 | 1.0e-4 | AdamW | 0.97 | Equal |
| RadNet-1.28M (0.1% data) | 256 | 1.0e-4 | AdamW | 0.97 | Equal |

**Table ST4: Hyperparameters.**
Additional settings can be found in the config files at our code repository (https://github.com/TIO-IKIM/Transfer-learning-across-domain-boundaries). The "decay factor" entry refers to learning rate decay; all training runs used a decaying learning rate, where the learning rate would be reduced by multiplying with the decay factor after each epoch. ImageNet and RadNet fine-tuning tasks trained for 100 epochs, all other tasks for 50 epochs.

[+] - Surprisingly, working with the usual batch size dependency for learning rates in LARS (which usually result in very large learning rates $\geq 1$) caused significant instability and degraded performance in early experiments when fine-tuning on RadNet. Consequently, we reduced the learning rate until this behavior disappeared.